%% file: main.tex
\documentclass{article}

\usepackage[authoryear,round,comma,sort&compress]{natbib}

\usepackage[preprint,nonatbib]{neurips_2025}

\usepackage[utf8]{inputenc} % allow utf-8 input
\usepackage[T1]{fontenc}    % use 8-bit T1 fonts
\usepackage{hyperref}       % hyperlinks
\usepackage{url}            % simple URL typesetting
\usepackage{amsfonts}       % blackboard math symbols
\usepackage{amsmath}        % extended math
\usepackage{bm}             % bold math symbols
\usepackage{microtype}      % microtypography
\usepackage{xcolor}         % colors
\usepackage{algorithm}      % algorithms
\usepackage{amssymb}            % Defines symbols like \mathbb{R}
\usepackage{mathtools}          % Extra math tools extending amsmath
\usepackage{mathrsfs}           % Enables \mathscr
\usepackage{graphicx}           % Include images
\usepackage{subcaption}         % Subfigures
\usepackage[space]{grffile}     % Allow spaces in image filenames

\usepackage{booktabs}           % Professional tables
\usepackage{algorithm}          % Algorithm environment
\usepackage{algpseudocode}      % Algorithm pseudocode
\usepackage{multirow}           % Table multirow cells
\usepackage{listings}           % Code blocks
\usepackage{wrapfig}            % Wrap text around figures
\definecolor{codegray}{rgb}{0.5,0.5,0.5}
\definecolor{codegreen}{rgb}{0.1,0.5,0.1}
\definecolor{codepurple}{rgb}{0.58,0,0.82}
\definecolor{backcolour}{rgb}{0.97,0.97,0.97}

\lstdefinestyle{rlcode}{
    language=Python,
    backgroundcolor=\color{backcolour},
    commentstyle=\color{codegreen},
    keywordstyle=\color{blue},
    stringstyle=\color{codepurple},
    numberstyle=\tiny\color{codegray},
    basicstyle=\ttfamily\tiny,
    breaklines=true,
    captionpos=b,
    keepspaces=true,
    numbers=left,
    numbersep=6pt,
    frame=single,
    tabsize=2,
    showstringspaces=false
}
\usepackage{float}
\lstdefinestyle{promptstyle}{
    basicstyle=\ttfamily\footnotesize,
    showstringspaces=false,
    breaklines=true,
}
\lstdefinestyle{pythonstyle}{
    language=Python,
    basicstyle=\ttfamily\small,
    keywordstyle=\color{blue},
    commentstyle=\color{gray},
    stringstyle=\color{teal},
    showstringspaces=false,
    breaklines=true,
    tabsize=4
}
\definecolor{commentpink}{RGB}{199,21,133}
\algrenewcommand\algorithmiccomment[1]{\hspace{1.5em}\textcolor{commentpink}{\textit{// #1}}}

\hypersetup{
    colorlinks=true,
    linkcolor=black,
    citecolor=blue,
    urlcolor=blue,
    filecolor=blue,
    pdfborder={0 0 0}
}

\input{math_commands.tex}

\title{Discovering Reinforcement Learning Interfaces with Large Language Models}

\author{%
  Akshat Singh Jaswal, Ashish Baghel, Paras Chopra \\
  Lossfunk \\
  \texttt{\{akshat.jaswal, ashish.baghel, paras\}@lossfunk.com}
}

\begin{document}

% Add logo at the very top of the document, before maketitle
\thispagestyle{empty}
\begin{center}
\vspace*{-1cm}
\includegraphics[width=0.25\textwidth]{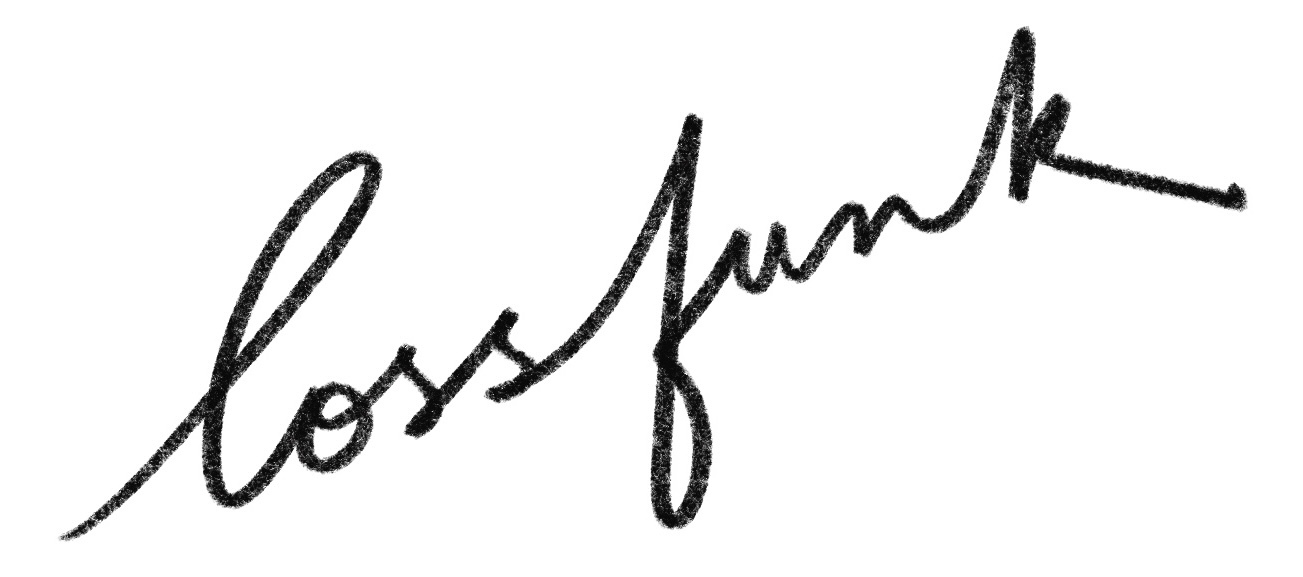}
\vspace{0.3cm}
\end{center}

\maketitle 

\begin{abstract}
Reinforcement learning systems rely on environment interfaces that specify observations and reward functions, yet constructing these interfaces for new tasks often requires substantial manual effort. While recent work has automated reward design using large language models (LLMs), these approaches assume fixed observations and do not address the broader challenge of synthesizing complete task interfaces. We study RL task interface discovery from raw simulator state, where both observation mappings and reward functions must be generated. We propose LIMEN\footnote{Code available at \url{https://github.com/Lossfunk/LIMEN}}, a LLM guided evolutionary framework that produces candidate interfaces as executable programs and iteratively refines them using policy training feedback. Across novel discrete gridworld tasks and continuous control domains spanning locomotion and manipulation, joint evolution of observations and rewards discovers effective interfaces given only a trajectory-level success metric, while optimizing either component alone fails on at least one domain. These results demonstrate that automatic construction of RL interfaces from raw state can substantially reduce manual engineering and that observation and reward components often benefit from co-design, as single-component optimization fails catastrophically on at least one domain in our evaluation suite.
\end{abstract}

% Key results summary - larger for better readability
\section{Introduction}

A central challenge in reinforcement learning is specifying the interface through which agents interact with environments, what they observe and how they are rewarded. While progress has been made in learning algorithms, the interface itself remains designed by human experts. Manually engineering these components is a critical bottleneck, as these design choices largely determine an agent's learning efficiency, exploration, and final policy performance \citep{sutton2018reinforcement}.

Recent work has explored automating reward design, often using large language models (LLMs) to generate reward functions or reward models from task descriptions or environment feedback \citep{llr1,llr2,llr3}. However, these approaches assume a fixed already tuned observation interface. In many environments, the raw state representations contain poorly structured information that can hinder learning, while carefully designed observations can substantially simplify the learning problem. Despite its importance, automatic interface discovery has been relatively underexplored compared to reward design.

\begin{figure}[ht]
    \centering
    \includegraphics[width=0.7\linewidth]{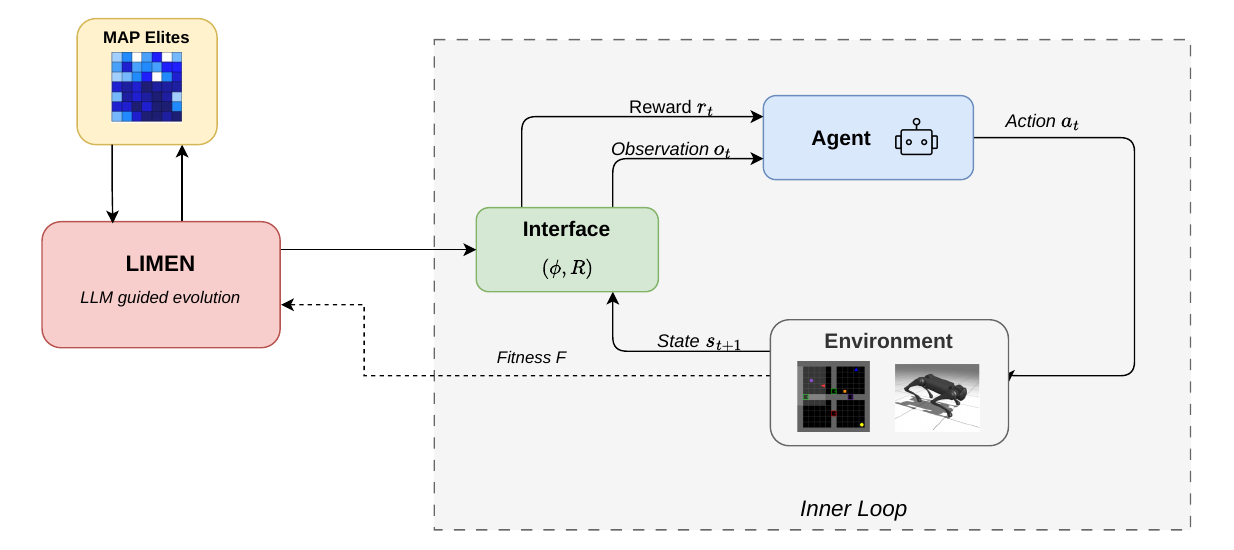}
    \caption{Overview of the LIMEN framework. The outer loop performs evolutionary search: LIMEN selects a parent interface from the MAP-Elites archive, mutates it via LLM-guided code generation, and evaluates the resulting interface by training an RL agent in the inner loop. The interface $(\phi, R)$ mediates between the raw simulator state and the agent, defining the observations and rewards that constitute the induced MDP. Fitness is measured by trajectory-level task success and fed back to update the archive.}
    \label{fig:overview}
\end{figure}

In this work, we address the problem of RL interface discovery by jointly optimizing the observation mapping and reward function. We formalize this interface as a pair $(\phi, R)$, where $\phi$ maps environment states to observations and $R$ specifies the reward function; together, these induce the effective Markov Decision Process (MDP) experienced by the agent. We assume access to a trajectory-level success metric that evaluates whether the task was completed for example, whether the agent reached the goal or maintained tracking error below a threshold. This metric serves as the fitness signal for evolutionary search but is distinct from the per-step reward function, which must be discovered.
Given this signal, we propose \textbf{LIMEN} (\textbf{L}earning \textbf{I}nterfaces via \textbf{M}DP-guided \textbf{E}volutio\textbf{N}), a method that discovers effective interfaces using LLM-guided mutation and evolutionary search. By representing $\phi$ and $R$ as executable programs, LIMEN evolves candidate interfaces through a quality-diversity archive, evaluating each by training RL agents to measure performance. Figure~\ref{fig:overview} illustrates the overall LIMEN framework. Starting from a task description and raw simulator state, the system generates candidate observation and reward programs, evaluates them through policy learning, and iteratively refines the interface using evolutionary selection.

\begin{figure}[t]
\centering
\includegraphics[width=0.65\textwidth]{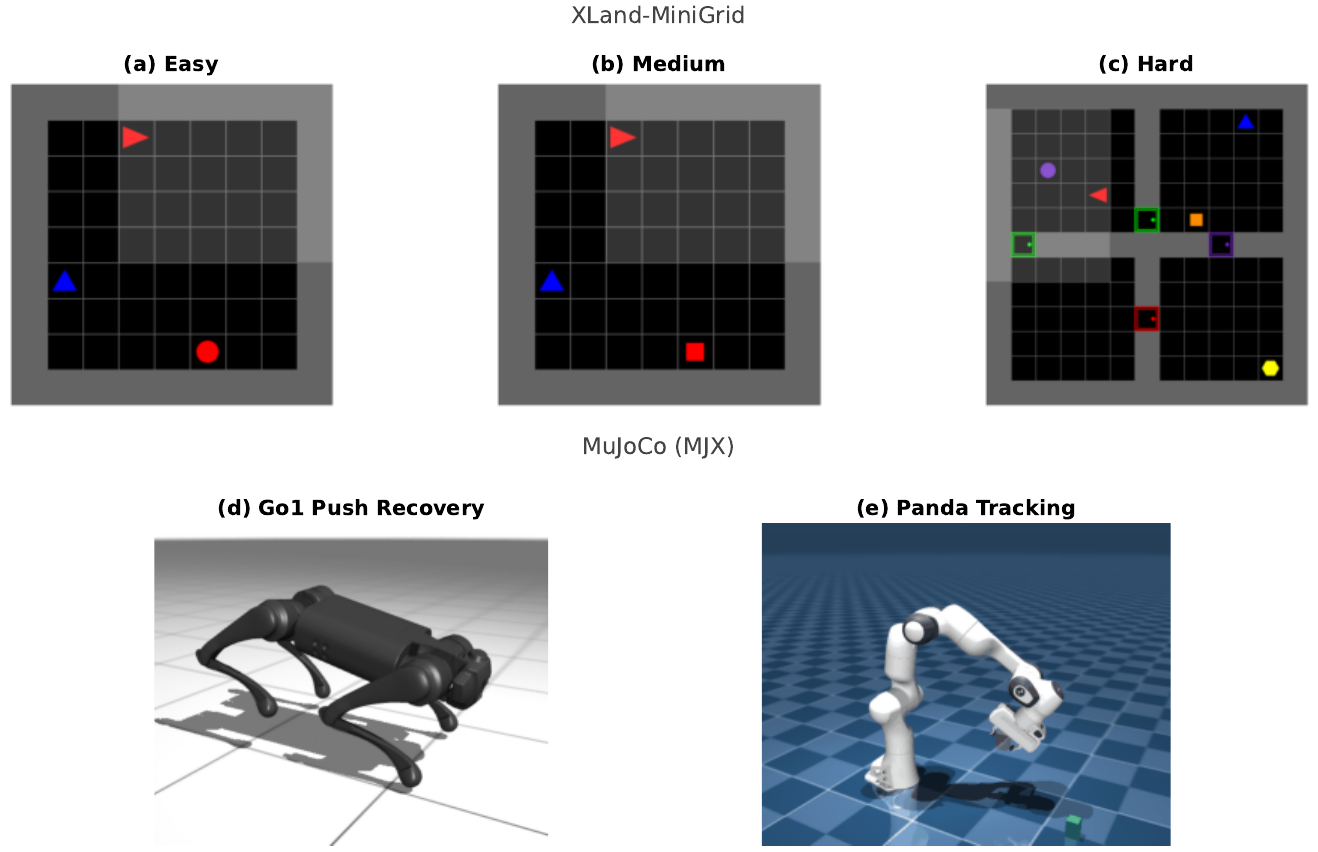}
\caption{\textbf{Evaluation environments.}
Top: XLand-MiniGrid tasks of increasing compositional complexity—
(a) object pickup among distractors,
(b) relational placement,
(c) multi-step rule chain across rooms.
Bottom: MuJoCo tasks—
(d) quadruped push recovery,
(e) manipulator trajectory tracking.}
\label{fig:environments}
\end{figure}

To evaluate this approach, we design a suite of novel experiments across gridworld reasoning tasks and robotic control environments. Figure~\ref{fig:environments} illustrates the environments used in our experiments.

These experiments specifically test the ability of \textbf{LIMEN} to generate effective interfaces for novel tasks, demonstrating that jointly evolving observations and rewards is the only approach that avoids catastrophic failure across all five tasks, whereas observation-only and reward-only optimization each fail on at least one domain. Analysis of the discovered interfaces further reveals consistent, interpretable patterns in both observation features and reward shaping strategies that emerge through the evolutionary process.

\section{Problem Formulation}

\subsection{RL Interface and Induced MDP}

We assume access to a simulator world model defined by a Markov decision process (MDP) \citep{puterman1994markov}

\[
\mathcal{M} = (\mathcal{S}, \mathcal{A}, T, \rho_0),
\]

where $\mathcal{S}$ is the simulator state space, $\mathcal{A}$ is the action space, 
$T(s' \mid s, a)$ denotes the transition dynamics, and $\rho_0$ is the initial state distribution.

We assume that a task-specific success metric 
$F : \Pi \rightarrow \mathbb{R}$ is available, 
which evaluates the performance of a trained policy $\pi$ over full episodes and serves as the fitness function.

An \textbf{RL interface} is defined as a pair

\[
\mathcal{I} = (\phi, R),
\]

where:
\begin{itemize}
\item $\phi : \mathcal{S} \rightarrow \mathcal{O}$ is an observation mapping,
\item $R : \mathcal{S} \times \mathcal{A} \times \mathcal{S} \rightarrow \mathbb{R}$ is a reward function.
\end{itemize}

The interface transforms the simulator into an induced learning problem.
Given $\mathcal{I}$, we define the induced MDP

\[
\mathcal{M}_{\mathcal{I}} = (\mathcal{O}, \mathcal{A}, T_\phi, R),
\]

where observations are given by $o_t = \phi(s_t)$ and $T_\phi$ denotes the observation-level dynamics induced by $T$.

\subsection{Interface Discovery Objective}

Let $\mathcal{A}_{\text{RL}}$ denote a reinforcement learning algorithm.
Given an interface $\mathcal{I} = (\phi, R)$, we denote by

\[
\pi_{\mathcal{I}} = \mathcal{A}_{\text{RL}}(\mathcal{M}_{\mathcal{I}})
\]

the policy obtained by training on the induced MDP.

We study the problem of task interface discovery from raw simulator state.
The objective is to identify an interface that maximizes task performance under the evaluation metric $F$:

\[
\mathcal{I}^* =
\arg\max_{\phi, R}
\; \mathbb{E}_{\xi}
\left[
F\big( \pi_{\mathcal{I}} \big)
\right],
\]

where the expectation is taken over sources of stochasticity $\xi$, including policy initialization, environment randomness, and training noise.

This defines a bilevel optimization problem:

\[
\begin{aligned}
\textbf{Outer level:} \quad
& \max_{\phi, R} \; F(\pi_{\phi,R}) \\
\textbf{Inner level:} \quad
& \pi_{\phi,R} = \mathcal{A}_{\text{RL}}(\mathcal{M}_{\phi,R})
\end{aligned}
\]

In this setting, the search space for $\phi$ and $R$ consists of executable programs that operate directly on the raw simulator state variables $\mathcal{S}$.

\section{Related Work}

\paragraph{Reward Optimization in Reinforcement Learning.}
A large body of work studies the construction of reward functions for reinforcement learning. Classical approaches include inverse reinforcement learning and imitation learning from demonstrations \citep{il1,il4,il5}, as well as preference-based and RLHF methods that infer reward models from human feedback \citep{rlhf1,rlhf2}. More recently, large language models have been used to automatically generate reward code from natural language task descriptions (e.g., Eureka, Text2Reward, DrEureka) \citep{llr1,llr2,llr3}. These approaches optimize the reward function within a fixed environment interface, assuming that the observation space is already sufficient for learning. As a result, these methods cannot address settings where learning fails due to missing or poorly structured observations. A related line of work in inverse reinforcement learning jointly learns observation models alongside reward functions from demonstrations \citep{arora2023online, levine2010feature, il3}, but does so within policy-learning pipelines that produce neural embeddings over a fixed input space. In contrast, we consider a strictly more general problem: searching over executable programs that define the induced MDP itself, jointly synthesizing observation mappings and reward functions from raw simulator state. The novelty of our formulation lies in framing this as explicit programmatic interface search, producing interpretable and transferable code artifacts rather than learned embeddings.

\paragraph{Representation Learning and State Abstraction.}
Representation learning has long been recognized as central to RL performance, with methods ranging from auxiliary losses and contrastive objectives to bisimulation metrics and state abstraction. These approaches learn neural representations jointly with the policy \citep{wang2024state,paischer2024semantic}. However, they do not alter the observation function provided to the agent rather they learn embeddings over a fixed input space. Our work instead searches over explicit, executable observation mappings that redefine the agent’s input space prior to learning. This allows the dimensionality, structure, and semantics of observations to change, effectively altering the learning problem rather than learning representations within it.

\paragraph{Evolutionary and Programmatic Search in RL.}
Evolutionary algorithms have long been combined with reinforcement learning for policy search, hyperparameter tuning, and hybrid optimization \citep{hao2023erlre2efficientevolutionaryreinforcement,pourchot2019cemrlcombiningevolutionarygradientbased,pmlr-v235-li24cp}. 
More recently, large language models have been integrated into evolutionary pipelines as structured mutation operators for program synthesis and reward evolution \citep{chen2023evoprompting, wei2025lero}. 
Beyond RL, systems such as OpenEvolve and AlphaEvolve demonstrate that LLM-guided evolutionary refinement can effectively search over executable program space by iteratively proposing, evaluating, and improving code \citep{novikov2025alphaevolvecodingagentscientific,openevolve}. 
Our work applies this paradigm to a different object, the reinforcement learning interface itself.
Rather than evolving policies or optimizing reward components alone, we use quality-diversity search to explore complete observation and reward programs that define the induced MDP faced by the agent.

To our knowledge, no prior work performs joint programmatic search over observation mappings and reward functions to automatically construct reinforcement learning interfaces from raw simulator state.
\section{Method}

We address RL interface discovery using LLM guided evolutionary search. Given a task description and environment specification,
the system synthesizes executable observation and reward space and optimizes them through iterative training feedback.

Each interface consists of two executable programs operating on simulator state: (1) an observation mapping producing agent inputs and (2) a reward function generating scalar rewards.

\paragraph{Interface Representation.}
Interfaces are represented as Python programs operating directly on the raw simulator state. 
Observation programs construct fixed-size feature vectors from environment state variables using JAX-compatible numerical operations (e.g., arithmetic transforms, concatenation, norms, and differentiable conditionals).
Reward programs compute scalar rewards from state transitions $(s_t, a_t, s_{t+1})$ and may utilize environment-provided statistics such as cumulative errors or episode progress. 
Observation dimensionality is constrained to a maximum of 512 features to ensure stable training.

\subsection{Evolutionary Interface Search}
We formulate the discovery of $\mathcal{I}$ as an iterative search over the space of Python programs. As detailed in Algorithm~\ref{alg:limen_system}, we utilize a MAP-Elites archive, a Quality-Diversity method \citep{pugh} to maintain a population of well performing and structurally diverse solutions. Each iteration can generate multiple parallel candidate interfaces but we chose to generate a single candidate for our experiments which is evaluated by training PPO agents across three random seeds to estimate its fitness.

\begin{algorithm}[t]
\caption{LIMEN: Learning Interfaces via MDP-guided Evolution}
\label{alg:limen_system}
\begin{algorithmic}[1]
\Require Task description $D$, RL algorithm $\mathcal{A}_{\text{RL}}$, fitness metric $F$, LLM $p_\theta$
\State Initialize MAP-Elites archive $\mathcal{A}$
\State $\mathcal{I}_0 \sim p_\theta(D)$ \Comment{Initial interface}
\State Evaluate $\mathcal{I}_0$ and insert into $\mathcal{A}$

\For{$i = 1,\dots,N$}
    \State $\mathcal{I}_p \leftarrow \Call{Select}{\mathcal{A}}$ \Comment{Parent selection}
    \State $\mathcal{I}_i \sim p_\theta(\Call{BuildPrompt}{D,\mathcal{I}_p})$ \Comment{LLM mutation}
    \If{$\Call{Validate}{\mathcal{I}_i}$} \Comment{Program validation}
        \State Train $\pi_i \leftarrow \mathcal{A}_{RL}(\mathcal{M}_{\mathcal{I}_i})$
        \State $f_i \leftarrow \mathbb{E}_s[F(\pi_i^{(s)})]$ \Comment{Mean success}
        \State $\Call{Insert}{\mathcal{A},\mathcal{I}_i,f_i,[\dim(\phi_i),AST(R_i)]}$
    \EndIf
\EndFor

\State \Return best interface in $\mathcal{A}$
\end{algorithmic}
\end{algorithm}

\textbf{Prompt Synthesis as Mutation.} The LLM $p_\theta$ acts as a structured mutation operator \citep{austin2021programsynthesislargelanguage}. For each iteration, a mutation prompt $\mathcal{P}$ is synthesized containing the task description $D$, a parent interface $\mathcal{I}$ sampled from the archive, and the top-performing interfaces from the archive and recently failed programs with their error traces. This feedback loop steers the LLM away from negative code patterns and toward robust implementations.
Prompt sections are randomly sampled and shuffled, and candidate programs are generated via stochastic decoding to encourage exploration.
\textbf{Program Validation and Safety.} Candidates undergo validation including syntax checks, dependency loading, and execution tests to ensure observation and reward outputs have valid shapes. Together with the short budget cascade filter and the island-model diversity pressure, these mechanisms filter out degenerate candidates such as those producing constant rewards or shape invalid observation vectors before they consume full training budget.

\subsection{Quality-Diversity Archive and Selection}

To maintain diversity and prevent the search from collapsing to a single interface strategy, we employ a MAP-Elites archive~\citep{mouret2015illuminatingsearchspacesmapping} structured by two behavioral descriptors: \textit{observation dimensionality} and \textit{reward structural complexity} (measured by Abstract Syntax Tree (AST) node count). Concretely, the archive is a 2D grid: Axis 1 bins observation dimensionality into uniform ranges (e.g., 1–50, 51–100, \ldots, 451–512), and Axis 2 bins reward AST node count similarly. These descriptors capture the primary structural axes along which interfaces vary: a compact 10-feature observation paired with a simple reward occupies a different niche than a 200-feature observation with complex multi-term shaping.

Following the standard island model technique in evolutionary computation~\citep{whitley1999island}, the archive is partitioned into $K$ independent islands that evolve in parallel, with the highest-fitness interface from each island migrating to its neighbor at fixed intervals. When selecting a parent for mutation, we sample from the global archive 70\% of the time (fitness-proportional) and from the local island 30\% of the time (uniform) to balance exploitation of strong interfaces with localized exploration although this ratio is adjustable. In practice, with 30 candidates evaluated per run, the archive remains sparse typically 15-20 occupied cells but the diversity pressure is sufficient to prevent repeated refinement of a single interface design and instead encourages structurally distinct solutions.

\subsection{Inner-Loop Evaluation and Fitness}
The fitness of an interface $F(\mathcal{I})$ is determined by the performance of an agent $\pi$ trained from scratch within the induced MDP $\mathcal{M}_{\mathcal{I}}$.

\textbf{Evaluation Cascade.} We also utilize a "short-budget" cascade filter: candidates must exceed a minimum success threshold which can be set up by the user in a truncated training run before proceeding to full multi-seed evaluation.

\textbf{Fitness Formulation.} $F(\mathcal{I})$ is defined as the mean success rate across evaluation seeds (e.g., goal acquisition in XLand or tracking precision in MuJoCo). While our experiments primarily utilize success-based metrics, the LIMEN framework is agnostic to the specific fitness signal, allowing for the integration of auxiliary objectives or domain-specific performance indicators without modifying the core discovery loop.

\section{Experiments}
\label{sec:experimental_setup}
We evaluate \textsc{LIMEN} on five tasks spanning discrete reasoning and continuous robotics control. Specifically, we examine whether joint interface discovery can automatically construct effective observation and reward functions from raw simulator state, whether optimizing these components jointly provides consistent benefits compared to optimizing them independently, and whether the resulting interfaces generalize beyond nominal training conditions.

\subsection{Environments}

\paragraph{XLand-MiniGrid.}
We evaluate three tasks built on XLand-MiniGrid~\citep{xland-minigrid}, a JAX-based gridworld library designed for compositional reasoning.

\textbf{Easy.} The agent must pick up a specified object in a $9 \times 9$ grid containing distractors (80-step horizon).

\textbf{Medium.} The agent must place one object adjacent to another specified object, introducing relational reasoning ($9 \times 9$, 80-step horizon).

\textbf{Hard.} A $13 \times 13$ multi-room environment with a 400-step horizon requiring a sequence of ordered subgoals.

Default observations expose a flattened $7 \times 7$ egocentric grid without explicit relational structure. The built-in reward is sparse ($+1$ on task completion, $0$ otherwise).

\paragraph{MuJoCo Robotics.}

We design two continuous-control tasks simulated using MuJoCo MJX ~\citep{mujoco}.

\textbf{Go1 Push Recovery.}
A Unitree Go1 quadruped must maintain balance for 500 simulation steps while subjected to random lateral force impulses (150–400\,N) applied every 75 steps. An episode succeeds if the robot survives the entire episode and maintains average base displacement below 10\,cm.

\textbf{Panda Tracking.}
A Franka Panda 7-DoF manipulator must track a moving 3D Lissajous trajectory (radius 0.10\,m, angular speed 0.35\,rad/s) for 500 steps. Success requires maintaining mean end-effector error below 2\,cm.

These domains present complementary challenges: gridworld tasks are primarily observation-limited, while robotics tasks are reward-sensitive due to sparse signals in high-dimensional control. Although evaluated on five tasks, the framework itself is environment agnostic, with RL training as the primary computational bottleneck mitigated through parallel JAX simulation. The full environment documentation provided to the LLM during interface generation is included in the Supplementary Material ~\ref{sup:envctx}.

\subsection{Training Configuration}

All agents are trained using Proximal Policy Optimization (PPO) \citep{schulman2017proximal}, using either the default XLand-MiniGrid implementation or the Brax PPO implementation \citep{brax} depending on the environment.
Across all experiments we keep the RL algorithm, architectures, and hyperparameters fixed to isolate the effect of interface design.
Each evolution run consists of 30 iterations, generating and evaluating one candidate interface per iteration. Full PPO hyperparameters, training budgets, and network architectures are provided in Supplementary Material ~\ref{sup:training}.
\subsection{Evolution Protocol}

Canditate interfaces are generated using Claude Sonnet 4.6 (temperature 0.7) and evaluated by training an RL agent from scratch. Full prompt templates and LLM configuration details used for interface generation are provided in the Supplementary Material ~\ref{sup:evol}.

For XLand-MiniGrid we employ cascade evaluation. A short training run filters candidates (Easy: 500K steps, Medium: 1M steps, Hard: 3M steps). Candidates exceeding a small success threshold (1–5\%) proceed to full multi-seed training (Easy: 1M steps, Medium: 2M steps, Hard: 5M steps) with three random seeds.

\begin{figure}[H]
\centering
\includegraphics[width=0.9\linewidth]{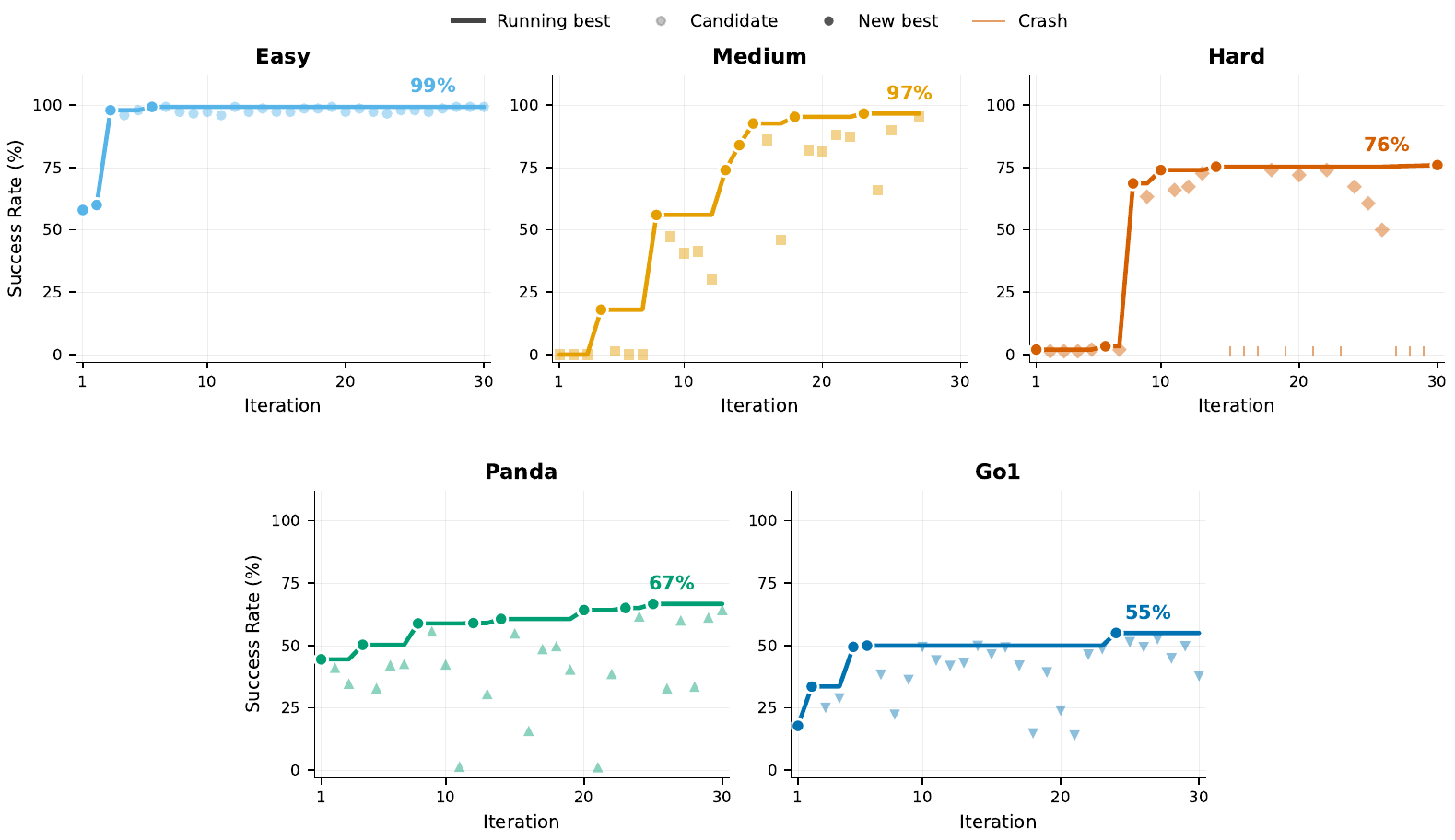}
\caption{Evolution progress of LIMEN showing candidate interfaces, crash events, and improvements in the running best success rate across iterations.}
\label{fig:evolution_progress}
\end{figure}

For MuJoCo tasks we skip cascade filtering and run full training directly (Panda: 15M steps, Go1: 25M steps) with three seeds. Fitness is defined as the mean success rate across seeds.

A full evolution run consists of 30 iterations. XLand runs require approximately 1–3 GPU hours, while MuJoCo runs require 6–7 hours. LLM cost per run is approximately \$3–11.
Figure~\ref{fig:evolution_progress} shows the evolution dynamics of LIMEN,
including candidate interfaces explored during search and improvements in
the best discovered success rate over iterations.

We report results from a single evolution run per task, 
additional runs across five seeds (Appendix~\ref{sup:seed_variance}) 
show reliable convergence on Easy and Medium, with higher 
variance on Hard.

\subsection{Baselines}

We compare joint interface discovery against three ablations:

\textbf{Sparse.} Raw simulator observations with binary success reward.

\textbf{Obs-Only.} Evolves the observation mapping while keeping the reward fixed.

\textbf{Reward-Only.} Evolves the reward function while keeping observations fixed to raw simulator state. This is a controlled instantiation of LLM-based reward search methods such as Eureka~\citep{llr1} and Text2Reward~\citep{llr2}.

All baselines use identical evolution budgets and RL training configurations.

\subsection{Main Results}

To eliminate post-selection bias from the evolutionary search, we retrain the best discovered interface for each method from scratch under fixed training budgets and evaluate performance over 10 independent seeds since evolution selects from ~30 candidates using noisy 3 seed estimates. 

\begin{figure}[H]
\centering
\includegraphics[width=0.9\linewidth]{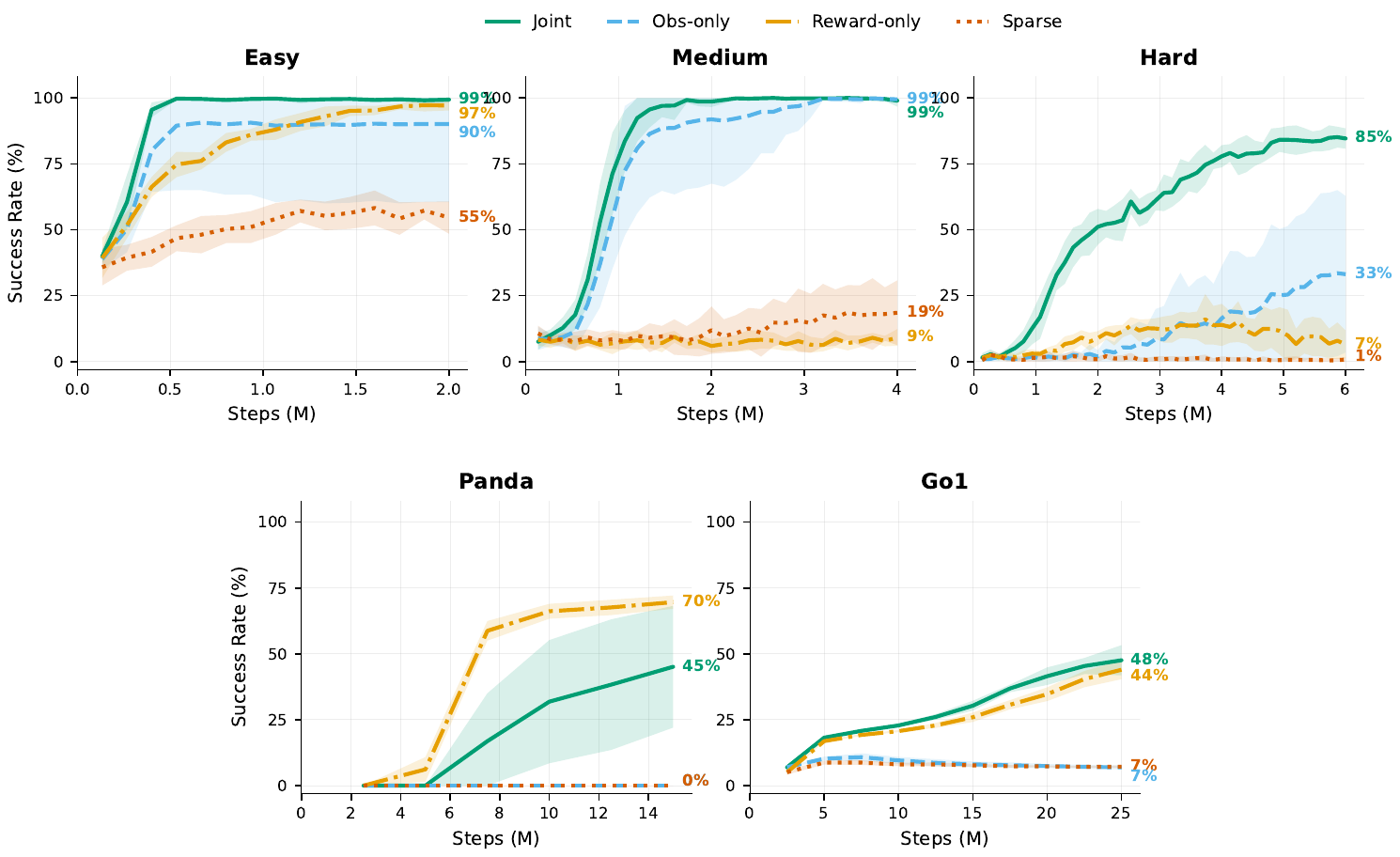}
\caption{Learning curves for LIMEN and ablations across five tasks. 
Success rate versus environment steps (millions), averaged over 10 seeds with shaded standard deviation. 
Joint interface discovery consistently achieves higher performance than observation-only, reward-only, and sparse baselines.}
\label{fig:learning_curves}
\end{figure}

Evaluation budgets are 2M steps for Easy, 4M for Medium, 6M for Hard, and 15M steps for Panda and 25M steps for Go1. Figure~\ref{fig:learning_curves} shows learning curves

  Joint discovery consistently achieves higher performance than observation only, reward only, and sparse baselines, reaching 99\% (Easy), 99\% (Medium), 85\% (Hard),
  45\% (Panda), and 48\% (Go1). The sparse baseline fails on all but the
  easiest task, confirming that raw interfaces are insufficient for
  complex domains.

  The ablations reveal complementary failure modes. Reward-only search
  collapses on Medium (19\%) and Hard (1\%), while observation-only
  search fails entirely on Panda (0\%). On individual tasks,
  single-component ablations can match or exceed joint optimization, observation-only
  reaches 100\% on Easy, and reward-only reaches 70\% on Panda but no
  single ablation succeeds broadly. Each fails catastrophically on at
  least one domain. We analyze why these bottlenecks arise in
  Section~\ref{sec:analysis}.

  All methods use the same fitness signal (mean success rate) during
  evolution for fair comparison. In practice, the LLM tends to construct unnecessarily large observation vectors when unconstrained (e.g., 174 features for Easy), penalizing observation dimensionality in the fitness function is a promising direction we leave to future work.
  \subsection{Independent LLM Sampling Baseline}                                   

\begin{figure}[H]
\centering
\includegraphics[width=0.7\linewidth]{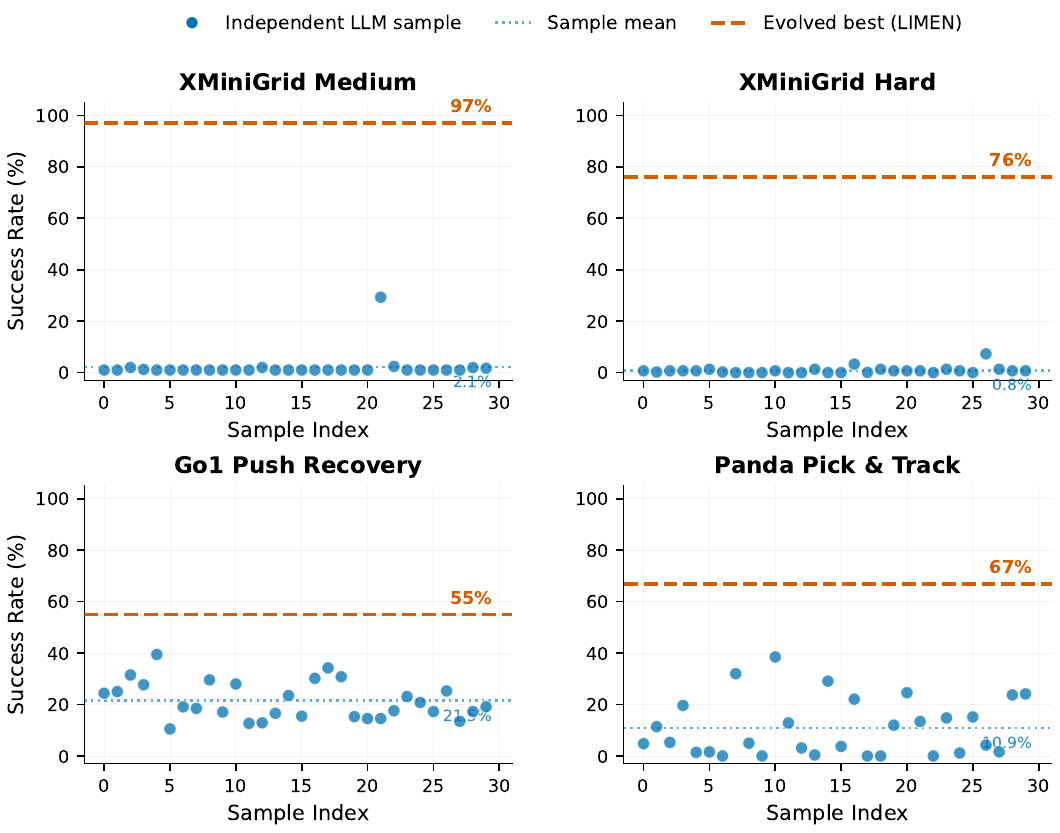}
\caption{Independent LLM samples (no evolution) versus the best interface found by LIMEN across four tasks. Each dot is a single interface sampled from the LLM with the same prompt and evaluated over 3 seeds with identical training budgets.}
\label{fig:no_evolution}
\end{figure}

  To isolate the contribution of the evolutionary loop, we evaluate a natural baseline: sampling interfaces independently
  from the LLM using the same task prompt, without iterative feedback or selection pressure. We draw 30 independent
  samples per task and train each under identical conditions (same RL algorithm, network architecture, and compute budget;
   3 seeds). Figure~\ref{fig:no_evolution} plots each sample's success rate alongside the best interface discovered by
  LIMEN.

  The gap is substantial across all environments. On XMiniGrid Medium and Hard, independent samples achieve mean success
  rates of $2.1\%$ and $0.8\%$ respectively, compared to $97\%$ and $76\%$ with evolution. On the robotics tasks, independent
  sampling reaches $21.5\%$ (Go1) and $10.9\%$ (Panda) on average, well below the evolved $55\%$ and $67\%$. Notably, even the
   best of 30 independent samples falls far short of the evolved interface in every case, indicating that the LLM's
   prior over interface designs is insufficient on its own, the iterative evaluate-and-refine loop is essential for
  navigating the combinatorial space of observation selections and reward compositions.
\subsection{Robustness to Distribution Shift}

\begin{wrapfigure}{r}{0.43\textwidth}
\vspace{-35pt}
\centering
\includegraphics[width=0.43\textwidth]{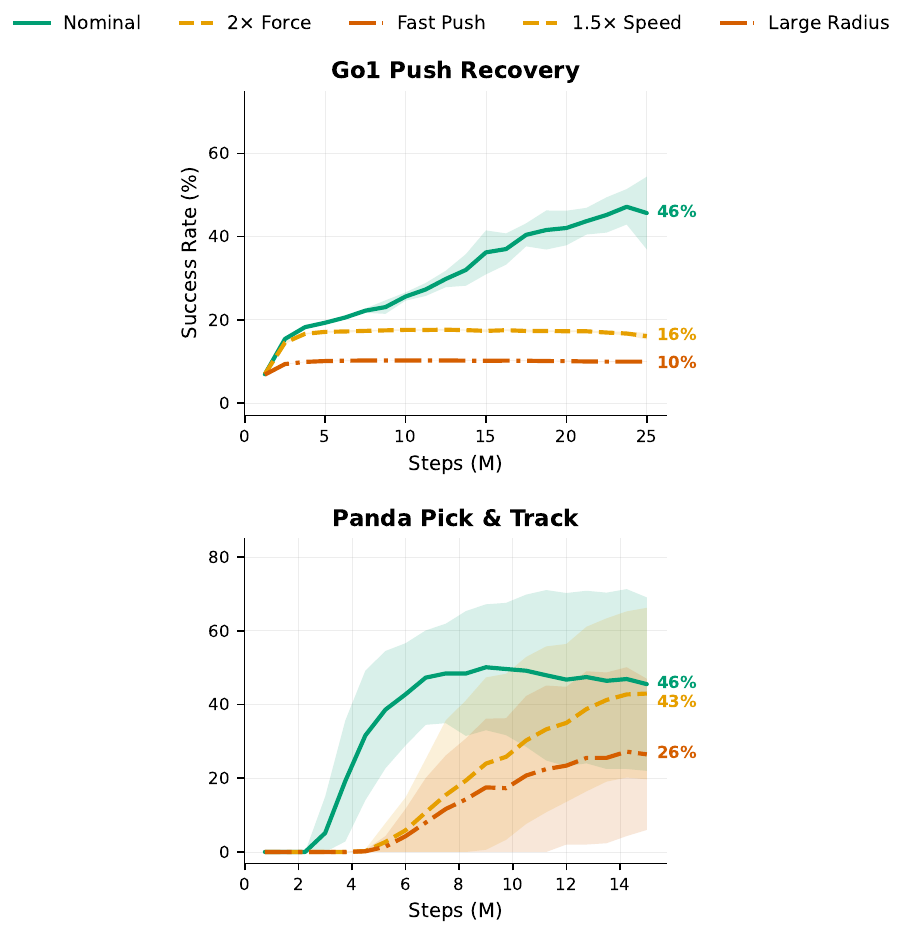}
\caption{Robustness under distribution shift for the Go1 push recovery and Panda tracking tasks.}
\label{fig:robustness}
\vspace{-10pt}
\end{wrapfigure}
  To evaluate robustness, we retrain the best robotics interfaces under perturbed dynamics (Go1: 25M steps, Panda: 15M    
  steps, 5 seeds each).                 
Performance degrades continuously under both perturbation types rather than collapsing to zero. For Go1, doubling push force reduces
   success from $50.3\%$ to $17.8\%$, while increasing push frequency reduces it to $10.3\%$. For Panda, increasing
  trajectory speed produces only modest degradation ($53.9\% \to 45.1\%$), while enlarging the tracking radius has a larger
  effect ($53.9\% \to 29.9\%$). These results suggest that evolved interfaces retain task relevant structure rather than collapsing completely under perturbation though the magnitude of degradation varies with perturbation type. The resulting learning dynamics are illustrated in Figure~\ref{fig:robustness}.
  
  Unlike conventional transfer learning, we transfer the task interface, what the agent observes and optimizes rather than a policy or representation. The fact that these interfaces continue to yield non-trivial learning under perturbed dynamics rather than collapsing entirely, suggests that the search captures some task-relevant structure beyond the specific nominal parameters.
\section{Analysis}
\label{sec:analysis}
  
To understand how LIMEN improves learning performance, we analyze the observation mappings and reward functions discovered during evolution. Despite the large program search space, several consistent structural patterns emerge.
Complete examples of evolved interface for each task are provided in the supplementary material ~\ref{sup:discoveredinterface}.

\subsection{Observation vs Reward Bottlenecks}

The ablation results reveal a clear distinction between tasks limited by observation design and those limited by reward shaping.

In the XLand-MiniGrid tasks, reward-only search fails despite well-structured rewards. For example, the evolved Medium reward includes phase-gated shaping and milestone bonuses yet achieves only 2\% success with the default observation space, while observation-only evolution reaches 99\%. Similar patterns appear in the Hard task, where reward-only search achieves only 4\% success. These results indicate that compositional gridworld tasks are primarily observation-limited, as the default observations lack structured relational information.  

The opposite pattern occurs in continuous-control domains. In the Panda tracking task, observation-only evolution fails entirely, while reward-only evolution achieves 71\% success with a simple four-term reward. This suggests that raw simulator observations already contain sufficient state information, but the sparse task reward provides insufficient learning signal. The observation only tends to outperform joint evolution in Panda due to the complexity of evolved observation space and we assume this could be mitigated by having a penalty for observation dimensionality in the fitness function. 

Together these results demonstrate that different domains fail for fundamentally different reasons, motivating joint optimization of observations and rewards.

\subsection{Recurring Interface Design Patterns}

Across tasks, LIMEN repeatedly discovers similar structural motifs.

Observation programs frequently construct relative geometric features such as position vectors, normalized distances, and directional indicators between task-relevant entities. Multi-scale encodings also appear frequently, along with explicit representations of task phase or progress.

Reward functions consistently incorporate potential-based shaping toward task goals, milestone bonuses for phase transitions, and smoothness penalties in continuous-control domains.

These structures closely resemble reward shaping and representation engineering strategies commonly designed manually by RL practitioners, suggesting that LIMEN rediscover many of the components that make RL problems easier to learn.

\subsection{Case Study: Structural Reward Discovery}

Evolution can also modify the structure of reward functions. In the Go1 push-recovery task, an early interface gates the position reward by uprightness, preventing position gradients until the agent learns to stand. A later interface removes this gating, providing continuous gradients encouraging recovery even when partially unstable.

This structural change improves success from 32\% to 55\% and introduces additional features such as multi-scale position encodings and body-frame coordinates. This example illustrates that LIMEN can discover qualitatively different reward structures that significantly alter learning dynamics.
\section{Limitations and Future Work}

LIMEN relies on an external evaluation metric that measures true task success and guides interface evolution. In domains where such a reliable metric is unavailable or difficult to specify, the evolution may become much harder. In addition, the primary practical bottleneck is computational cost. Our experiments mitigate this cost using JAX based environments that enable highly parallel simulation, but evaluation cost grows with RL training. Evaluating the approach on larger-scale environments and tasks with high-dimensional observations such as vision remains an important direction for future work. Our current evolutionary search is deliberately simple, single candidate iterations with a basic MAP-Elites archive. More sophisticated evolutionary strategies could improve search efficiency within the same compute budget. Similarly, using more capable frontier models as the mutation operator may yield higher-quality candidates per iteration, reducing the number of iterations required while increasing the number of iterations could also yield better results.

Finally, our current formulation assumes access to structured simulator state variables when constructing observation programs, which effectively provides privileged information not always available in real world settings. However, many simulation environments expose structured state variables during development, and such privileged information is commonly used for reward design and debugging in RL research. Developing scalable evaluation strategies and reducing reliance on privileged simulator state are promising directions for enabling interface discovery in more complex real world environments.

\section{Conclusion}

This work studies the problem of reinforcement learning interface discovery, where both observation mappings and reward functions must be automatically constructed from raw simulator state given only a trajectory-level success metric. We introduce LIMEN, an LLM-guided evolutionary framework that searches over executable interface programs and evaluates them through policy learning. Across gridworld reasoning and robotics control tasks, our experiments show that learning fails for fundamentally different reasons: compositional tasks are primarily observation-limited, while continuous control tasks are frequently reward-limited, and single-component optimization fails catastrophically on at least one domain. These results suggest that observations and rewards often benefit from co-design, and that jointly optimizing them can substantially reduce the manual effort required to formulate effective reinforcement learning problems.

% References section with proper linking

% Bibliography - using author-year style compatible with natbib
\bibliographystyle{apalike}
\bibliography{references}

% Include appendix with detailed voting method explanations
\input{appendix}

\end{document}

%% file: math_commands.tex
\def\eqref#1{equation~\ref{#1}}
\def\1{\bm{1}}

\DeclareMathAlphabet{\mathsfit}{\encodingdefault}{\sfdefault}{m}{sl}
\SetMathAlphabet{\mathsfit}{bold}{\encodingdefault}{\sfdefault}{bx}{n}

%% file: appendix.tex
% APPENDIX - ICLR 2026
% The Sequential Edge: Supplementary Materials

\clearpage
\appendix
\section{Training Details}
\label{sup:training}
\subsection{PPO Hyperparameters}

\subsubsection{XMinigrid (Discrete --- RNN-PPO)}

\begin{table}[H]
\centering
\caption{PPO hyperparameters for XMinigrid environments.}
\begin{tabular}{lccc}
\toprule
Hyperparameter & Easy & Medium & Hard \\
\midrule
Learning rate & $10^{-3}$ & $10^{-3}$ & $10^{-3}$ \\
LR schedule & linear decay & linear decay & linear decay \\
Discount ($\gamma$) & 0.99 & 0.99 & 0.99 \\
GAE ($\lambda$) & 0.95 & 0.95 & 0.95 \\
Clip ratio ($\epsilon$) & 0.2 & 0.2 & 0.2 \\
Entropy coefficient & 0.01 & 0.01 & 0.01 \\
Value coefficient & 0.5 & 0.5 & 0.5 \\
Max grad norm & 0.5 & 0.5 & 0.5 \\
Num environments & 8192 & 8192 & 8192 \\
Rollout length & 16 & 16 & 16 \\
Update epochs & 1 & 1 & 1 \\
Minibatches & 16 & 16 & 16 \\
Eval episodes & 80 & 80 & 80 \\
Fitness eval episodes & 50 & 50 & 50 \\
\bottomrule
\end{tabular}
\end{table}

\subsubsection{MuJoCo (Continuous --- Brax PPO)}

\begin{table}[H]
\centering
\caption{PPO hyperparameters for MuJoCo tasks.}
\begin{tabular}{lcc}
\toprule
Hyperparameter & Panda & Go1 \\
\midrule
Learning rate & $5\times10^{-4}$ & $3\times10^{-4}$ \\
Discount ($\gamma$) & 0.97 & 0.97 \\
GAE ($\lambda$) & 0.95 & 0.95 \\
Clip ratio ($\epsilon$) & 0.2 & 0.2 \\
Entropy coefficient & 0.015 & 0.01 \\
Value coefficient & 0.5 & 0.5 \\
Max grad norm & 1.0 & 1.0 \\
Num environments & 2048 & 4096 \\
Unroll length & 10 & 20 \\
Batch size & 1024 & 256 \\
Updates per batch & 8 & 4 \\
Num evaluations & 20 & 20 \\
Episode length & 500 & 500 \\
Normalize observations & Yes & Yes \\
Reward scaling & 1.0 & 1.0 \\
\bottomrule
\end{tabular}
\end{table}

\subsubsection{Training Budgets}

\begin{table}[H]
\centering
\caption{Training budgets in environment timesteps.}
\begin{tabular}{lccc}
\toprule
Task & Cascade (short) & Full & 10-seed Eval \\
\midrule
Easy & 500K & 1M & 2M \\
Medium & 1M & 2M & 4M \\
Hard & 3M & 5M & 6M \\
Panda & --- & 15M & 15M \\
Go1 & --- & 25M & 25M \\
\bottomrule
\end{tabular}
\end{table}

\subsection{Network Architectures}

\subsubsection{XMinigrid: RNN-PPO with GRU Memory}

The policy uses a recurrent PPO architecture with a GRU memory module.

\textbf{Architecture}

\begin{itemize}
\item Observation encoder: Dense(256) $\rightarrow$ ReLU $\rightarrow$ Dense(256)
\item Action embedding: Embedding($|\mathcal{A}|$, 16)
\item Previous reward: scalar input
\item Concatenated representation passed to GRU
\item GRU hidden size: 512
\end{itemize}

\textbf{Policy head}
\begin{itemize}
\item Dense(256) $\rightarrow$ Tanh $\rightarrow$ Dense($|\mathcal{A}|$)
\item Output distribution: categorical
\end{itemize}

\textbf{Value head}
\begin{itemize}
\item Dense(256) $\rightarrow$ Tanh $\rightarrow$ Dense(1)
\end{itemize}

\subsubsection{MuJoCo: Brax MLP-PPO}

\textbf{Panda}

\begin{itemize}
\item Policy network: MLP(32, 32, 32, 32)
\item Value network: MLP(256, 256, 256, 256, 256)
\item Observation normalization enabled
\end{itemize}

\textbf{Go1}

\begin{itemize}
\item Policy network: MLP(512, 256, 128)
\item Value network: MLP(512, 256, 128)
\item Observation normalization enabled
\end{itemize}
\section{Evolution Details}
\label{sup:evol}
\subsection{LLM Configuration}

\begin{table}[H]
\centering
\begin{tabular}{lcc}
\toprule
Parameter & XMinigrid & MuJoCo \\
\midrule
Model & Claude Sonnet 4.6 & Claude Sonnet 4.6 \\
Temperature & 0.7 & 0.7 \\
Max tokens & 64,000 & 128,000 \\
Retries & 3 & 3 \\
\bottomrule
\end{tabular}
\end{table}

\subsection{Prompt Templates}

Candidate interfaces are generated using a structured prompt consisting of a system prompt and a user prompt. The system prompt defines the design principles and output constraints, while the user prompt provides the task specification and evolutionary feedback.

\subsubsection{System Prompt}

\begin{lstlisting}[style=promptstyle]
You are an expert reinforcement learning engineer. You design
MDP interfaces for RL agents in reinforcement learning environments.

Your task is to write two JAX-compatible Python functions:
1. get_observation(state)
2. compute_reward(state, action, next_state)

The RL agent sees ONLY what get_observation returns and is trained
ONLY on what compute_reward provides.

Design philosophy:
- The goal is to help a neural network learn, not to solve the task.
- Observations must contain all information needed for learning.
- Normalize features to similar ranges.
- Observation size can vary (max 512 elements).

Reward design:
- Provide a useful learning signal guiding the agent toward the goal.
- The agent is evaluated on task success rate, not reward magnitude.
- Avoid reward hacking and misleading proxies.

Generalization:
- Environments are randomized across episodes.
- Do not hardcode environment-specific constants.
- Compute all features from the current state.

Output format:
Return code inside a single Python markdown fence.
Only `import jax` and `import jax.numpy as jnp` are allowed.
\end{lstlisting}

\subsubsection{From-Scratch User Prompt}

The first generation is created without a parent program.

\begin{lstlisting}[style=promptstyle]
## Task
{task_description}

## Instructions
Design get_observation(state) and compute_reward(state, action, next_state)
functions that allow a PPO agent to learn the task.

Observation design:
- Include all information required to solve the task.
- Compute features from the environment state.
- Normalize values to stable ranges.

Reward design:
- Provide a learning signal aligned with task completion.
- Reward real progress toward the goal.

Constraints:
- get_observation returns a 1D jnp.float32 vector
- compute_reward returns a scalar jnp.float32
- Only jax and jax.numpy imports are allowed
\end{lstlisting}

\subsubsection{Evolutionary User Prompt}

Subsequent generations improve on an existing program.

\begin{lstlisting}[style=promptstyle]
## Task
{task_description}

## Parent Program
--- BEGIN CODE ---
{parent_code}
--- END CODE ---
Success rate: {success_rate}
Reward: {final_return}
Observation dimension: {obs_dim}

## Feedback
- Recent failures
- Best programs discovered so far
- Diverse programs from different MAP-Elites cells
- Training feedback (variance, plateau detection)

## Instructions
Based on the feedback above, write an improved interface.

Constraints:
- Observation must return a 1D float32 vector
- Reward must return a scalar float32
- Only jax and jax.numpy imports allowed

Output code inside a single Python markdown fence.
\end{lstlisting}

\subsection{Improvement Guidance}

The evolutionary prompt includes targeted guidance depending on the
performance of the parent program.

\begin{table}[H]
\centering
\begin{tabular}{lc}
\toprule
Success Rate & Guidance Focus \\
\midrule
0\% & structural reset, redesign observation or reward \\
0--30\% & missing features or weak reward signal \\
30--60\% & analyze failure cases and enrich representation \\
60--90\% & targeted improvements and reward refinement \\
90\%+ & robustness and simplification \\
\bottomrule
\end{tabular}
\end{table}

\subsection{Evolution Seed Variance}
\label{sup:seed_variance}

\begin{figure}[H]
\centering
\includegraphics[width=0.9\linewidth]{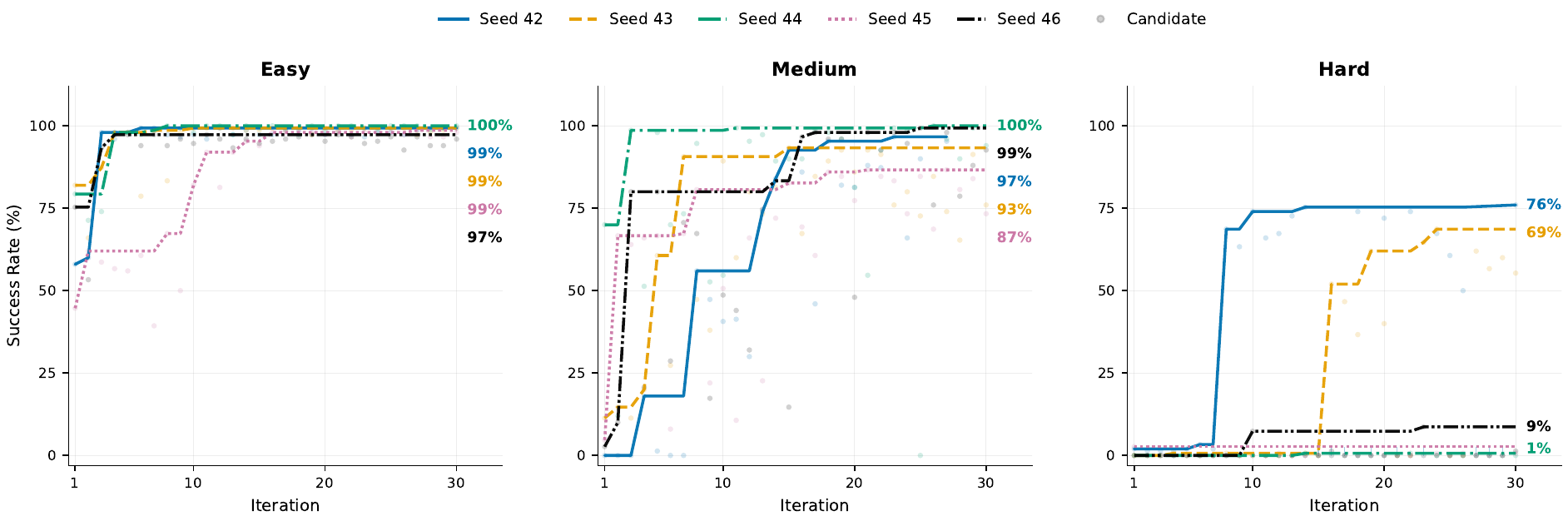}
\caption{Evolution progress across five independent seeds per XLand-MiniGrid task. Each line shows the running best success rate over 30 iterations; faded dots show individual candidate evaluations.}
\label{fig:evolution_seeds}
\end{figure}

To assess the reproducibility of evolutionary search, we repeat the XLand-MiniGrid evolution protocol across five independent seeds (42--46). Figure~\ref{fig:evolution_seeds} shows the running best success rate for each seed.

On Easy and Medium, all five seeds converge to high-performing interfaces (97--100\% and 87--100\% respectively), indicating reliable search on tractable tasks. On Hard, two seeds discover strong interfaces (76\% and 69\%) while three stall below 10\%, reflecting the combinatorial difficulty of jointly discovering multi-stage reward shaping and structured relational observations in large environments. Main paper results report Seed 42. Improving search reliability on complex tasks through longer evolution, better selection strategies, or stronger base models is a natural direction for future work.

\section{Environment Context Provided to the LLM}
\label{sup:envctx}
During interface discovery, the LLM is provided with environment-specific
documentation describing the task, available state fields, action spaces,
and implementation constraints. This documentation serves as the API
reference that the model uses when constructing observation and reward
functions.

\subsection{XLand-MiniGrid Context}

The following documentation was provided to the LLM when generating
interfaces for XLand-MiniGrid tasks.

\subsubsection{Task Environment}

XLand-MiniGrid environments consist of a grid world containing objects,
doors, and keys. The agent receives a partially structured environment
state and must interact using discrete actions.

The state object contains:

\begin{itemize}
\item \texttt{state.grid}: grid tensor of shape $(H, W, 2)$ where the first
channel encodes tile type and the second channel encodes color.
\item \texttt{state.agent.position}: agent position $(y, x)$.
\item \texttt{state.agent.direction}: agent orientation
($0=\text{up}, 1=\text{right}, 2=\text{down}, 3=\text{left}$).
\item \texttt{state.agent.pocket}: currently held object.
\item \texttt{state.step\_num}: current timestep.
\end{itemize}

The action space consists of six discrete actions:

\begin{itemize}
\item \texttt{FORWARD}
\item \texttt{TURN\_RIGHT}
\item \texttt{TURN\_LEFT}
\item \texttt{PICKUP}
\item \texttt{PUTDOWN}
\item \texttt{TOGGLE}
\end{itemize}

The LLM is instructed to implement two functions:

\begin{lstlisting}[style=pythonstyle]
def get_observation(state) -> jnp.ndarray:
    ...

def compute_reward(state, action, next_state) -> jnp.ndarray:
    ...
\end{lstlisting}

All implementations must be JAX-compatible.

\subsection{Go1 Push Recovery Context}

For the quadruped push-recovery task, the LLM receives a structured
description of the robot state and task objectives.

\subsubsection{Task}

The Unitree Go1 robot must remain upright while recovering from
random horizontal pushes applied to the torso.

The agent must:

\begin{itemize}
\item maintain upright orientation,
\item return to the origin after pushes,
\item maintain its heading direction.
\end{itemize}

An episode terminates if the robot tilts too far:

\[
\texttt{upvector}_z < 0.3
\]

Success requires surviving the full episode and maintaining an average
position error below $10\,\text{cm}$.

\subsubsection{State Information}

Key state fields available to the interface include:

\begin{itemize}
\item \texttt{state.data.qpos}: robot position and joint angles
\item \texttt{state.data.qvel}: linear and angular velocities
\item \texttt{state.data.actuator\_force}: actuator torques
\item \texttt{state.info["gyro"]}: IMU angular velocity
\item \texttt{state.info["gravity"]}: gravity vector in body frame
\item \texttt{state.info["local\_linvel"]}: body-frame linear velocity
\item \texttt{state.info["pos\_xy"]}: displacement from origin
\item \texttt{state.info["heading"]}: robot yaw angle
\item \texttt{state.info["push\_force"]}: current push impulse
\end{itemize}

The action space consists of 12 continuous joint offsets corresponding
to the quadruped's leg joints.

\subsection{Panda Tracking Context}

For the Panda arm tracking task, the LLM receives documentation describing
the target trajectory and available robot state.

\subsubsection{Task}

The Panda robot must track a moving target following a Lissajous
trajectory in three-dimensional space.

Success is defined as maintaining a mean end-effector tracking error
below $2\,\text{cm}$ over a 500-step episode.

\subsubsection{Target Trajectory}

The target follows a parametric Lissajous curve:

\[
x = 0.45 + 0.10 \sin(\omega_x t + \phi_x)
\]

\[
y = 0.00 + 0.10 \sin(\omega_y t + \phi_y)
\]

\[
z = 0.18 + 0.07 \sin(\omega_z t)
\]

where frequencies and phases are randomized at the start of each
episode.

\subsubsection{Available State Fields}

The LLM receives access to the following fields through
\texttt{state.info}:

\begin{itemize}
\item \texttt{target\_pos}: current target position
\item \texttt{target\_vel}: analytical velocity of the trajectory
\item \texttt{gripper\_pos}: end-effector position
\item \texttt{gripper\_target\_dist}: Euclidean tracking error
\item \texttt{traj\_params}: trajectory frequencies and phases
\item \texttt{prev\_ctrl}: previous control signal
\end{itemize}

The action space consists of continuous joint position deltas for
the seven arm joints and the gripper actuator.

\section{Discovered Interfaces}
\label{sup:discoveredinterface}

\subsection{XMinigrid Easy}

\textbf{Task.} Pick up a blue pyramid in a $9\times9$ grid within $80$ steps.

\textbf{Performance.} Success rate: \textbf{99\%}. Observation dimension: \textbf{174}.

\subsubsection{Observation}

The observation encodes the agent pose, pyramid location, and a local egocentric map.  
Object positions are represented using normalized coordinates and relative offsets to the agent.

\begin{table}[H]
\centering
\caption{Observation features for the Easy task.}
\begin{tabular}{l c}
\hline
Feature group & Dim \\
\hline
Agent position $(x,y)$ & 2 \\
Agent orientation (one-hot) & 4 \\
Pocket state (tile, color, flags) & 4 \\
Blue pyramid position & 2 \\
Relative offset to pyramid & 2 \\
Distance to pyramid & 1 \\
Directional indicators & 4 \\
Front tile features & 2 \\
Step progress & 1 \\
Local tile IDs ($7\times7$) & 49 \\
Local tile colors ($7\times7$) & 49 \\
Local pyramid indicator ($7\times7$) & 49 \\
\hline
Total & 174 \\
\hline
\end{tabular}
\end{table}

The $7\times7$ egocentric grid provides spatial context around the agent while the directional indicators help disambiguate target direction.

\subsubsection{Reward}

The reward combines a sparse pickup signal with dense shaping encouraging approach and correct orientation toward the pyramid.

\begin{table}[H]
\centering
\caption{Reward components for the Easy task.}
\begin{tabular}{l l}
\hline
Component & Definition \\
\hline
Pickup reward & $10\,\mathbf{1}_{pickup}$ \\
Approach shaping & $0.5(d_t-d_{t+1})$ \\
Adjacency bonus & $0.5\,\mathbf{1}_{adjacent}$ \\
Facing bonus & $1.0\,\mathbf{1}_{facing}$ \\
Step penalty & $-0.005$ \\
\hline
\end{tabular}
\end{table}
The reward structure combines a sparse task completion signal with dense geometric shaping based on the Manhattan distance to the target. The adjacency and facing bonuses encourage the agent to approach the pyramid from a valid pickup configuration, reducing exploration difficulty in the final interaction step. Together these terms transform the sparse pickup objective into a smooth navigation problem while preserving the correct optimal policy.

 \subsubsection{Interface Snippet}                                                                                       
  \begin{lstlisting}[style=pythonstyle]                                                                                   
  def get_observation(state):                                                                                             
      agent_y = state.agent.position[0] / (H - 1)         # (1)
      agent_x = state.agent.position[1] / (W - 1)         # (1)                                                           
      dir_oh = jax.nn.one_hot(state.agent.direction, 4)    # (4)                                                          
      pocket_tile = state.agent.pocket[0] / 12.0           # (1)
      pocket_color = state.agent.pocket[1] / 11.0          # (1)
      holding = (pocket[0] == PYRAMID) & (pocket[1] == BLUE)

      # Find blue pyramid via grid mask scan
      bp_mask = (grid[:,:,0] == PYRAMID) & (grid[:,:,1] == BLUE)
      bp_y = jnp.sum(yy * bp_mask) / jnp.maximum(count, 1)
      bp_x = jnp.sum(xx * bp_mask) / jnp.maximum(count, 1)
      rel_y = (bp_y - agent_y_raw) / H                     # (1)
      rel_x = (bp_x - agent_x_raw) / W                     # (1)
      dist = (abs(agent_y - bp_y) + abs(agent_x - bp_x))   # (1) Manhattan

      # Directional indicators
      bp_is_up    = (bp_y < agent_y)                        # (1)
      bp_is_down  = (bp_y > agent_y)                        # (1)
      bp_is_left  = (bp_x < agent_x)                        # (1)
      bp_is_right = (bp_x > agent_x)                        # (1)

      # 7x7 egocentric grid (tile IDs, colors, pyramid indicator)
      rows = jnp.clip(agent_row + offsets, 0, H - 1)
      local_tiles = grid[rows, cols, 0] / 12.0             # (49)
      local_colors = grid[rows, cols, 1] / 11.0            # (49)
      local_is_bp = (tiles == PYRAMID) & (colors == BLUE)  # (49)
      ...                                        # 174 total
      return jnp.concatenate([agent_pos, dir_oh, pocket,
          bp_pos, rel_offset, dist, directional_indicators,
          local_tiles, local_colors, local_is_bp])

  def compute_reward(state, action, next_state):
      just_picked_up = holding_now & (~was_holding)
      dist_before = abs(agent_y - bp_y) + abs(agent_x - bp_x)
      dist_after = abs(next_agent_y - bp_y) + abs(next_agent_x - bp_x)
      became_adjacent = (dist_after <= 1) & (dist_before > 1)
      facing_pyramid = (front_tile == PYRAMID) & (front_color == BLUE)

      reward = (10.0 * just_picked_up                    # task completion
              + 0.5 * (dist_before - dist_after)         # approach shaping
              + 0.5 * became_adjacent                    # adjacency milestone
              + 1.0 * facing_pyramid                     # orientation bonus
              - 0.005)                                   # step penalty
      return reward
  \end{lstlisting}
\subsection{XMinigrid Medium}

\textbf{Task.} Pick up a yellow pyramid and place it adjacent to a green square in a $9\times9$ grid within $80$ steps.

\textbf{Performance.} Success rate: \textbf{97\%}. Observation dimension: \textbf{102}.

\subsubsection{Observation}

The observation represents both task objects, their spatial relationships, and candidate placement locations around the square.

\begin{table}[H]
\centering
\caption{Observation features for the Medium task.}
\begin{tabular}{l c}
\hline
Feature group & Dim \\
\hline
Agent position and orientation & 6 \\
Inventory state & 2 \\
Yellow pyramid position & 3 \\
Green square position & 2 \\
Agent–pyramid offset & 2 \\
Agent–square offset & 2 \\
Pyramid–square offset & 2 \\
Pairwise distances & 3 \\
Front tile features & 7 \\
Adjacency indicators & 3 \\
Task phase indicators & 3 \\
Step progress & 1 \\
Best adjacent placement cell & 5 \\
Neighbor analysis around square & 8 \\
Local grid view ($5\times5$) & 50 \\
\hline
Total & 102 \\
\hline
\end{tabular}
\end{table}

In addition to relational features, the interface exposes the nearest valid floor cell adjacent to the square, which simplifies the placement planning problem.

\subsubsection{Reward}

The reward includes milestone rewards for pickup and correct placement along with shaping signals for approaching the pyramid and the square.

\begin{table}[H]
\centering
\caption{Reward components for the Medium task.}
\begin{tabular}{l l}
\hline
Component & Definition \\
\hline
Success reward & $10\,\mathbf{1}_{success}$ \\
Pickup reward & $2\,\mathbf{1}_{pickup}$ \\
Approach pyramid & $2(d_t^{ap}-d_{t+1}^{ap})$ \\
Approach square (holding) & $3(d_t^{as}-d_{t+1}^{as})$ \\
Incorrect placement penalty & $-1.5\,\mathbf{1}_{bad}$ \\
Correct placement bonus & $3\,\mathbf{1}_{correct}$ \\
Adjacency bonus & $2\,\mathbf{1}_{adjacent}$ \\
Putdown readiness bonus & $0.5\,\mathbf{1}_{ready}$ \\
\hline
\end{tabular}
\end{table}
The reward decomposes the task into two phases: locating the pyramid and placing it near the square. Distance-based shaping guides navigation during both phases, while adjacency and placement bonuses encourage correct positioning before executing the putdown action. By exposing placement structure through both the observation interface and the reward shaping, the agent can learn the sequential nature of the task more efficiently.

  \subsubsection{Interface Snippet}
  \begin{lstlisting}[style=pythonstyle]
  def get_observation(state):
      agent_pos = [agent_y / (H-1), agent_x / (W-1)]      # (2)
      dir_oh = jax.nn.one_hot(state.agent.direction, 4)    # (4)
      holding_yp = (pocket[0] == PYRAMID) & (pocket[1] == YELLOW)

      # Yellow pyramid & green square via mask scan
      yp_mask = (grid[:,:,0] == PYRAMID) & (grid[:,:,1] == YELLOW)
      gs_mask = (grid[:,:,0] == SQUARE) & (grid[:,:,1] == GREEN)
      yp_y = jnp.sum(yy * yp_mask) / jnp.maximum(count, 1)
      gs_y = jnp.sum(yy * gs_mask) / jnp.maximum(count, 1)

      # Pairwise spatial relations
      rel_agent_yp = (yp_pos - agent_pos) / H              # (2)
      rel_agent_gs = (gs_pos - agent_pos) / H              # (2)
      rel_yp_gs = (gs_pos - yp_pos) / H                   # (2)
      dists = [dist_agent_yp, dist_agent_gs, dist_yp_gs]  # (3)

      # Task phase indicators
      phase_pickup = (~holding_yp) & (~yp_adj_gs)          # (1)
      phase_carry = holding_yp                              # (1)
      phase_done = yp_adj_gs & (~holding_yp)               # (1)

      # Best adjacent floor cell to green square (loop over 4 dirs)
      for d in range(4):
          ny = clip(gs_y + DIRECTIONS[d, 0], 0, H-1)
          is_floor = (grid[ny, nx, 0] == FLOOR)
          better = is_floor & (dist < best_dist)
          best_adj = jnp.where(better, [ny, nx], best_adj)  # (5)

      front_ideal_putdown = front_is_floor * front_adj_to_gs
      local_grid = grid[agent-2:agent+3, :, :2] / [12, 11] # (50)
      ...                                        # 102 total
      return jnp.concatenate([agent_pos, dir_oh, obj_pos,
          pairwise_rels, dists, phases, best_adj, local_grid])

  def compute_reward(state, action, next_state):
      just_picked_up = (~was_holding) & now_holding
      just_succeeded = (~was_success) & now_adjacent

      reward = (10.0 * just_succeeded                    # task success
              + 2.0 * just_picked_up                     # pickup milestone
              + 2.0 * delta_to_pyramid * (~holding)      # phase 1: approach
              + 3.0 * delta_to_square * holding          # phase 2: deliver
              - 1.5 * placed_wrong_spot                  # wrong putdown
              + 3.0 * placed_adjacent                    # correct putdown
              + 2.0 * just_became_adj_to_square          # adjacency milestone
              + 0.5 * ready_to_putdown)                  # facing valid spot
      return reward
  \end{lstlisting}
\subsection{XMinigrid Hard}

\textbf{Task.} Pick up a blue pyramid (which transforms into a green ball when held) and place the ball adjacent to a yellow hex in a $13\times13$ four-room environment with distractors.

\textbf{Performance.} Success rate: \textbf{76\%}. Observation dimension: \textbf{147}.

\subsubsection{Observation}

The observation captures object localization, placement planning signals, and spatial context needed to solve the multi-stage task.

\begin{table}[H]
\centering
\caption{Observation features for the Hard task.}
\begin{tabular}{l c}
\hline
Feature group & Dim \\
\hline
Agent position and orientation & 6 \\
Pocket state & 4 \\
Blue pyramid localization & 6 \\
Yellow hex localization & 6 \\
Agent adjacency indicator & 1 \\
Front tile features & 8 \\
Agent neighbor analysis & 24 \\
Hex neighbor analysis & 20 \\
Best placement cell features & 4 \\
Green ball tracking & 3 \\
Task phase indicators & 3 \\
Phase-conditioned target vector & 3 \\
Step progress & 1 \\
Local grid view ($5\times5$) & 50 \\
Valid putdown indicators & 8 \\
\hline
Total & 147 \\
\hline
\end{tabular}
\end{table}

The interface exposes candidate placement cells adjacent to the hex and dynamically switches the navigation target depending on whether the agent is holding the ball.

\subsubsection{Reward}

The reward structure includes milestone rewards and dense shaping for each phase of the task.

\begin{table}[H]
\centering
\caption{Reward components for the Hard task.}
\begin{tabular}{l l}
\hline
Component & Definition \\
\hline
Pickup milestone & $5\,\mathbf{1}_{pickup}$ \\
Success reward & $20\,\mathbf{1}_{success}$ \\
Incorrect placement penalty & $-5\,\mathbf{1}_{wrong}$ \\
Approach pyramid & $3(d_t^{ap}-d_{t+1}^{ap})$ \\
Approach hex & $3(d_t^{ah}-d_{t+1}^{ah})$ \\
Placement guidance & $1.5(d_t^{best}-d_{t+1}^{best})$ \\
First adjacency bonus & $3\,\mathbf{1}_{adjacency}$ \\
Putdown readiness bonus & $0.5\,\mathbf{1}_{ready}$ \\
Step penalty & $-0.01$ \\
\hline
\end{tabular}
\end{table}
The reward reflects the two-stage structure of the task: first acquiring the pyramid and then transporting the resulting ball to the hex. Phase-specific shaping terms guide navigation toward the active objective, while additional bonuses encourage reaching valid placement configurations before executing the final action. This decomposition significantly reduces exploration difficulty in the large four-room environment while maintaining a sparse success criterion.

  \subsubsection{Interface Snippet}
  \begin{lstlisting}[style=pythonstyle]
  def get_observation(state):
      agent_pos = [agent_y / (H-1), agent_x / (W-1)]      # (2)
      dir_oh = jax.nn.one_hot(state.agent.direction, 4)    # (4)
      holding_gb = (pocket[0] == BALL) & (pocket[1] == GREEN)

      # Blue pyramid & yellow hex via mask scan
      bp_mask = (grid[:,:,0] == PYRAMID) & (grid[:,:,1] == BLUE)
      yh_mask = (grid[:,:,0] == HEX) & (grid[:,:,1] == YELLOW)

      # 4 neighbors of agent: [tile, color, is_pyr, is_hex,
      #                         is_floor, adj_to_hex] per dir
      for d in range(4):                                    # (24)
          ...

      # 4 neighbors of hex: [is_floor, rel_y, rel_x,
      #                       dist_to_agent, at_this_cell]
      for d in range(4):                                    # (20)
          ...

      # Best placement: closest floor tile adjacent to hex
      for d in range(4):
          is_floor = (grid[hex_y+dy, hex_x+dx, 0] == FLOOR)
          best = jax.lax.select(is_floor & closer, ...)    # (4)

      # Phase-dependent navigation target
      phase = [searching, holding, placed]                  # (3)
      target = jax.lax.select(holding > 0.5,
                   hex_pos, pyramid_pos)                    # (3)
      local_grid = grid[agent-2:agent+3, :, :2] / [12, 11] # (50)

      # Valid putdown per direction
      for d in range(4):                                    # (8)
          valid = is_floor * adj_to_hex
      ...                                        # 147 total
      return jnp.concatenate([agent_pos, dir_oh, pocket,
          bp_loc, hex_loc, neighbors, hex_neighbors,
          best_placement, phase, target, local_grid, ...])

  def compute_reward(state, action, next_state):
      just_picked_up = (~was_holding) & now_holding_green_ball
      ball_placed = was_holding & (~now_holding)
      success = ball_placed & ball_adj_to_hex

      reward = (20.0 * success                           # task completion
              + 5.0 * just_picked_up                     # pickup milestone
              - 5.0 * ball_placed & (~ball_adj_to_hex)   # wrong placement
              + 3.0 * delta_to_pyramid * (~holding)      # phase 1 shaping
              + 3.0 * delta_to_hex * holding             # phase 2 shaping
              + 1.5 * delta_to_best_placement * holding  # placement guidance
              + 3.0 * just_reached_hex_adjacency         # adjacency milestone
              + 0.5 * ready_to_putdown                   # facing valid spot
              - 0.01)                                    # step penalty
      return reward
  \end{lstlisting}
\subsection{GO1 Push Recovery}

\textbf{Task.} The quadruped must remain standing near the origin while random impulses are applied to the torso. An episode is successful if the robot survives the full duration and the average position error remains below $10$ cm.

\textbf{Performance.} Success rate: \textbf{55\%}. Observation dimension: \textbf{98}.

\subsubsection{Observation}

The observation combines orientation and velocity signals with joint state, control history, and disturbance information. In addition to raw proprioceptive signals, the interface includes features describing the robot’s displacement from the origin, projected future motion, and directional velocities toward the goal location.

\begin{table}[H]
\centering
\caption{Observation features for the GO1 push-recovery task.}
\begin{tabular}{l c}
\hline
Feature group & Dim \\
\hline
Gravity vector (body frame) & 3 \\
Up vector (world frame) & 3 \\
Angular velocity (gyro) & 3 \\
Linear velocity (body frame) & 3 \\
Linear velocity (world frame) & 3 \\
Position offset from origin (coarse) & 2 \\
Position offset from origin (fine) & 2 \\
Body-frame position offset & 2 \\
Body-frame position offset (fine) & 2 \\
Distance to origin & 1 \\
Direction to origin (world frame) & 2 \\
Direction to origin (body frame) & 2 \\
Height feature & 1 \\
Heading representation (sin/cos) & 2 \\
Joint angle deviations & 12 \\
Joint velocities & 12 \\
Previous action & 12 \\
External push force & 3 \\
Push activity indicator & 1 \\
Velocity toward origin (world) & 1 \\
Velocity toward origin (body) & 1 \\
Velocity magnitude & 1 \\
Uprightness indicator & 1 \\
Tilt danger indicator & 1 \\
Gyro magnitude & 1 \\
Base orientation quaternion & 4 \\
Actuator forces & 12 \\
Projected future position & 2 \\
Projected future distance & 1 \\
Horizontal angular velocity & 2 \\
\hline
Total & 98 \\
\hline
\end{tabular}
\end{table}

Several features explicitly encode recovery-relevant quantities such as tilt danger, velocity toward the origin, and projected future position, enabling the policy to anticipate destabilizing motion after external pushes.

\subsubsection{Reward}

The reward encourages the robot to remain upright while staying near the origin and smoothly recovering from disturbances. Dense shaping terms provide gradients for both stabilization and position correction.

\begin{table}[H]
\centering
\caption{Reward components for the GO1 task.}
\begin{tabular}{l l}
\hline
Component & Definition \\
\hline
Uprightness reward & $4\,u_z$ \\
Upright bonus & $10\max(0,u_z-0.8)^2$ \\
Tilt penalty & $-20\max(0,0.65-u_z)^2$ \\
Position reward & $4\,r_{pos}(d)$ \\
Progress toward origin & $2(d_t-d_{t+1})$ \\
Success proximity bonus & $2e^{-20d}$ \\
Heading stabilization & $0.5e^{-3|\theta|}$ \\
Height reward & $0.3\tanh(7h)$ \\
Velocity guidance & $r_{vel}$ \\
Angular velocity penalty & $-r_{gyro}$ \\
Action smoothness penalty & $-r_{smooth}$ \\
Torque penalty & $-r_{torque}$ \\
Fall penalty & $-20\,\mathbf{1}_{fall}$ \\
Survival bonus & $0.3$ \\
\hline
\end{tabular}
\end{table}

Here $u_z$ denotes the vertical component of the up vector and $d$ the distance from the origin.

The reward design balances stabilization and goal tracking. Uprightness remains the dominant signal for preventing falls, while multi-scale position rewards and progress shaping encourage the robot to return to the origin after disturbances. Velocity-dependent terms guide corrective motion when far from the target but discourage unnecessary movement once the robot has stabilized near the origin. Together these signals produce robust push recovery behavior while maintaining energy-efficient and smooth control.

  \subsubsection{Interface Snippet}
  \begin{lstlisting}[style=pythonstyle]
  DEFAULT_POSE = jnp.array([0.1, 0.9, -1.8, ...])  # x4 legs

  def get_observation(state):
      gravity    = state.info["gravity"]                   # (3) body-frame
      gyro       = state.info["gyro"] / 5.0                # (3)
      lin_vel    = state.info["local_linvel"] / 3.0        # (3) body-frame

      # Multi-scale position encoding
      pos_xy     = state.info["pos_xy"]                    # (2)
      pos_dist   = jnp.linalg.norm(pos_xy)
      pos_coarse = pos_xy / 1.0                            # (2) broad gradient
      pos_fine   = jnp.clip(pos_xy / 0.1, -5, 5)          # (2) sharp at origin

      # Direction to origin rotated into body frame
      heading    = state.info["heading"]
      dir_world  = -pos_xy / (pos_dist + 1e-6)            # (2)
      dir_body   = jnp.array([                             # (2) body-frame
          cos(-heading)*dir_world[0] - sin(-heading)*dir_world[1],
          sin(-heading)*dir_world[0] + cos(-heading)*dir_world[1]])

      # Stability indicators
      up_z        = state.info["upvector"][-1]
      tilt_danger = jnp.maximum(0.0, 0.7 - up_z)          # (1)

      # Predictive: linear extrapolation 0.2s ahead
      future_pos  = pos_xy + state.data.qvel[:2] * 0.2    # (2)

      joint_dev   = (qpos[7:] - DEFAULT_POSE) / pi         # (12)
      joint_vel   = qvel[6:] / 15.0                        # (12)
      push_force  = state.info["push_force"] / 400.0       # (3)
      push_active = jnp.tanh(push_mag / 100.0)             # (1)
      actuator_force = state.data.actuator_force / 50.0    # (12)
      ...                                        # 98 total
      return jnp.concatenate([gravity, gyro, lin_vel,
          pos_coarse, pos_fine, dir_body, tilt_danger,
          joint_dev, joint_vel, push_force, future_pos, ...])

  def compute_reward(state, action, next_state):
      up_z      = next_state.info["upvector"][-1]
      pos_dist  = jnp.linalg.norm(next_state.info["pos_xy"])
      prev_dist = jnp.linalg.norm(state.info["pos_xy"])

      # Uprightness (primary survival signal)
      upright   = 4.0 * up_z
      upright_b = 10.0 * jnp.maximum(0, up_z - 0.8)**2
      tilt_pen  = -20.0 * jnp.maximum(0, 0.65 - up_z)**2

      # Multi-scale position (NOT gated by uprightness)
      position  = 4.0 * (0.3*exp(-pos_dist)
                        + 0.4*exp(-5*pos_dist)
                        + 0.3*exp(-20*pos_dist))

      # Adaptive velocity: toward origin when far, still when near
      far = jnp.tanh(pos_dist * 8.0)
      vel_toward = jnp.dot(vel_xy, -pos_xy / (pos_dist+1e-6))
      vel_reward = far * jnp.tanh(vel_toward * 3) * 0.6

      progress  = 2.0 * clip((prev_dist - pos_dist)*40, -1, 1)
      fall_pen  = -20.0 * (up_z < 0.3)
      survival  = 0.3
      ...                                        # 14 terms total
      return upright + upright_b + tilt_pen + position
           + progress + vel_reward + fall_pen + survival + ...
  \end{lstlisting}

\subsection{Panda Target Tracking}

\textbf{Task.} The Panda arm must track a moving 3D target following a Lissajous trajectory using the end-effector. An episode is successful if the mean tracking error remains below $2$\,cm over the full trajectory.

\textbf{Performance.} Success rate: \textbf{67\%}. Observation dimension: \textbf{94}.

\subsubsection{Observation}

The observation encodes robot proprioception together with task-specific tracking information. In addition to joint state and control history, the interface exposes multi-scale tracking errors, target motion derivatives, and predictive features describing future target positions along the trajectory.

\begin{table}[H]
\centering
\caption{Observation features for the Panda tracking task.}
\begin{tabular}{l c}
\hline
Feature group & Dim \\
\hline
Joint positions (normalized) & 7 \\
Joint velocities & 7 \\
Gripper position & 3 \\
Target position & 3 \\
Tracking error (fine / medium / coarse) & 9 \\
Distance to target (multi-scale) & 3 \\
Target velocity & 3 \\
Target acceleration & 3 \\
Target jerk & 3 \\
Previous control input & 7 \\
Control–joint error & 7 \\
Trajectory frequency parameters & 3 \\
Initial trajectory phase & 4 \\
Current trajectory phase & 6 \\
Future tracking errors (multi-horizon) & 18 \\
Future target positions & 6 \\
Velocity–error alignment & 1 \\
Normalized time step & 1 \\
\hline
Total & 94 \\
\hline
\end{tabular}
\end{table}

Predictive features such as future target positions and trajectory phase information allow the policy to anticipate the motion of the target rather than reacting purely to instantaneous tracking error.

\subsubsection{Reward}

The reward is designed to provide a smooth signal for accurate tracking while encouraging stable and energy-efficient control.

\begin{table}[H]
\centering
\caption{Reward components for the Panda tracking task.}
\begin{tabular}{l l}
\hline
Component & Definition \\
\hline
Primary tracking reward & $0.6\,e^{-\frac{1}{2}(d/0.02)^2} + 0.3\,e^{-\frac{1}{2}(d/0.05)^2} + 0.1\,e^{-\frac{1}{2}(d/0.15)^2}$ \\
Velocity alignment bonus & $0.05\,a_{vel}$ \\
Control change penalty & $-0.002\,\|\Delta u\|$ \\
Action magnitude penalty & $-0.001\,\|u\|$ \\
\hline
\end{tabular}
\end{table}

Here $d$ denotes the end-effector distance to the target and $a_{vel}$ measures alignment between the tracking error direction and the target velocity.

The reward emphasizes precise tracking through a multi-scale Gaussian objective that strongly rewards errors below the $2$\,cm success threshold while still providing gradients when the robot is farther away from the target. Velocity alignment encourages the end-effector to move coherently with the target trajectory, improving dynamic tracking performance. Small control penalties regularize the policy and promote smooth arm motion without dominating the primary tracking objective.

  \subsubsection{Interface Snippet}
  \begin{lstlisting}[style=pythonstyle]
  def get_observation(state):
      arm_qpos   = state.data.qpos[0:7] / jnp.pi          # (7) normalized
      arm_qvel   = state.data.qvel[0:7] / 2.0              # (7)
      gripper    = state.info["gripper_pos"]                # (3)
      target     = state.info["target_pos"]                 # (3)
      error      = target - gripper

      # Multi-scale error normalization
      error_fine   = error / 0.02                           # (3) 2cm scale
      error_med    = error / 0.05                           # (3) 5cm scale
      error_coarse = error / 0.15                           # (3) 15cm scale
      dist_feats = [dist/0.02, dist/0.05, dist/0.15]      # (3)

      # Target dynamics: analytical derivatives of Lissajous
      target_vel  = state.info["target_vel"] / 0.15         # (3)
      target_acc  = -A * w**2 * sin(w*t + phi) / 0.03      # (3)
      target_jerk = -A * w**3 * cos(w*t + phi) / 0.05      # (3)

      # Trajectory phase encoding (sin/cos per axis)
      phase_x = [sin(w_x*t + phi_x), cos(w_x*t + phi_x)]  # (2)
      phase_y = [sin(w_y*t + phi_y), cos(w_y*t + phi_y)]   # (2)
      phase_z = [sin(w_z*t), cos(w_z*t)]                   # (2)

      # Multi-horizon future tracking errors
      for dt in [0.04, 0.10, 0.20, 0.40, 0.80, 1.60]:     # (18)
          future_target = compute_lissajous(t + dt)
          future_err = clip((future_target - gripper) / s, -5, 5)

      ctrl_joint_err = (prev_ctrl - arm_qpos) / 0.3        # (7)
      vel_error_align = dot(target_vel, error_dir)          # (1)
      ...                                        # 94 total
      return jnp.concatenate([arm_qpos, arm_qvel,
          error_fine, error_med, error_coarse, dist_feats,
          target_vel, target_acc, target_jerk,
          phase_x, phase_y, phase_z, future_errs, ...])

  def compute_reward(state, action, next_state):
      dist = next_state.info["gripper_target_dist"]

      # Multi-scale Gaussian: precise near 2cm threshold
      tight  = exp(-0.5 * (dist/0.02)**2)
      medium = exp(-0.5 * (dist/0.05)**2)
      coarse = exp(-0.5 * (dist/0.15)**2)
      tracking = 0.6*tight + 0.3*medium + 0.1*coarse

      # Velocity alignment: reward moving with target
      error_dir = error / (norm(error) + 1e-6)
      dist_weight = clip(dist / 0.05, 0, 1)
      vel_bonus = 0.05 * dot(error_dir, target_vel_dir) * dist_weight

      ctrl_pen = -0.002 * norm(ctrl_change)
      act_pen  = -0.001 * norm(action[:7])
      return tracking + vel_bonus + ctrl_pen + act_pen
  \end{lstlisting}

\section{Environment Adapter Architecture}
\label{sup:adapter}

To support multiple environment families with a unified discovery pipeline,
we implement an \textit{environment adapter} abstraction. Each environment
implements a small protocol that exposes environment construction,
success evaluation, and training routines.

\subsection{Adapter Interface}

\begin{lstlisting}[style=pythonstyle]
class EnvAdapter(ABC):
    """Abstract protocol for environment integration."""

    def make_env(self, get_obs_fn, reward_fn):
        """Create environment with injected MDP interface.

        Args:
            get_obs_fn: (state) -> observation vector
            reward_fn:  (prev_state, action, state) -> scalar reward
        Returns:
            (env, env_params)
        """

    def get_dummy_state(self):
        """Return a dummy state for crash-filter dry runs."""

    def compute_success(self, rollout_stats, env_params):
        """Compute task success metric."""

    def get_default_obs_fn(self):
        """Raw observation baseline (no feature engineering)."""

    def get_default_reward_fn(self):
        """Sparse binary reward baseline."""

    def run_training(self, config, interface, obs_dim, total_timesteps):
        """Train a PPO agent with the given interface."""
        env, params = self.make_env(
            interface.get_observation,
            interface.compute_reward
        )
        # PPO training loop
        return metrics

    def run_training_multi_seed(self, config, interface,
                                obs_dim, total_timesteps, num_seeds=3):
        """Train across multiple seeds."""
        results = [self.run_training(..., seed=s) for s in range(num_seeds)]
        return average_metrics(results)
\end{lstlisting}

Concrete adapters implement this interface for each environment family
(e.g., XLand-MiniGrid and MuJoCo/Brax environments).

\subsection{MDP Interface Injection}

The evolved observation and reward functions are injected into the
environment via a wrapper that intercepts environment transitions.

\begin{lstlisting}[style=pythonstyle]
class MDPInterfaceWrapper(Wrapper):
    """Inject evolved observation and reward functions."""

    def __init__(self, env, get_obs_fn=None, reward_fn=None):
        self._get_obs = get_obs_fn
        self._reward = reward_fn

    def reset(self, params, key):
        timestep = self.env.reset(params, key)

        if self._get_obs:
            timestep = timestep.replace(
                observation=self._get_obs(timestep.state))

        return timestep

    def step(self, params, timestep, action):
        prev_state = timestep.state
        next_timestep = self.env.step(params, timestep, action)

        if self._get_obs:
            next_timestep = next_timestep.replace(
                observation=self._get_obs(next_timestep.state))

        if self._reward:
            next_timestep = next_timestep.replace(
                reward=self._reward(prev_state, action,
                                    next_timestep.state))

        return next_timestep
\end{lstlisting}

This wrapper transparently replaces the environment's default
observation and reward functions with the evolved interface
without modifying the underlying simulator.

\section{Runtime and Cost}
\label{sup:cost}
All experiments were executed on a single NVIDIA L4 GPU. Each evolution run
consists of 30 iterations, evaluating one candidate interface per iteration.
We report wall-clock runtime and LLM usage statistics for both the main
experiments and ablation runs.
  \subsection{Main Evolution Runs}                                                                                        
   
  \begin{table}[H]                                                                                                        
  \centering      
  \caption{Runtime and LLM cost for main discovery runs.}
  \begin{tabular}{lcccc}
  \toprule
  Task & Iters & Wall Time (hr) & Tokens (M) & LLM Cost \\
  \midrule
  Easy   & 30 & 0.43 & 0.42 & \$1.94 \\
  Medium & 30 & 1.05 & 0.64 & \$2.85 \\
  Hard   & 30 & 0.82 & 1.15 & \$5.61 \\
  Panda  & 30 & 3.57 & 0.75 & \$3.42 \\
  Go1    & 30 & 5.56 & 0.80 & \$3.68 \\
  \midrule
  \textbf{Total} & \textbf{150} & \textbf{11.43} & \textbf{3.76} & \textbf{\$17.50} \\
  \bottomrule
  \end{tabular}
  \end{table}

  \subsection{Ablation Experiments}

  \begin{table}[H]
  \centering
  \caption{Runtime and LLM usage for ablation experiments.}
  \begin{tabular}{lcccc}
  \toprule
  Task & Wall Time (hr) & Tokens (M) & LLM Cost \\
  \midrule
  Easy (obs-only)        & 0.78 & 0.30 & \$1.33 \\
  Easy (reward-only)     & 0.35 & 0.22 & \$0.99 \\
  Medium (obs-only)      & 0.49 & 0.61 & \$2.68 \\
  Medium (reward-only)   & 0.16 & 0.28 & \$1.24 \\
  Hard (obs-only)        & 0.59 & 0.66 & \$2.91 \\
  Hard (reward-only)     & 0.53 & 0.36 & \$1.61 \\
  Panda (obs-only)       & 4.76 & 0.82 & \$3.79 \\
  Panda (reward-only)    & 4.76 & 0.82 & \$3.79 \\
  Go1 (obs-only)         & 5.94 & 0.66 & \$3.03 \\
  Go1 (reward-only)      & 6.31 & 0.70 & \$3.22 \\
  \midrule
  \textbf{Total}         & \textbf{24.68} & \textbf{5.43} & \textbf{\$24.60} \\
  \bottomrule
  \end{tabular}
  \end{table}

  \subsection{Aggregate Cost}

  \begin{table}[H]
  \centering
  \caption{Total compute and LLM cost across all experiments.}
  \begin{tabular}{lcccc}
  \toprule
  Experiment Group & Runs & Wall Time (hr) & Tokens (M) & LLM Cost \\
  \midrule
  Main experiments & 5  & 11.43 & 3.76 & \$17.50 \\
  Ablation experiments & 10 & 24.68 & 5.43 & \$24.60 \\
  \midrule
  \textbf{Total} & \textbf{15} & \textbf{36.11} & \textbf{9.20} & \textbf{\$42.10} \\
  \bottomrule
  \end{tabular}
  \end{table}

  Across all experiments, LIMEN consumed approximately
  \textbf{9.2M LLM tokens} and \textbf{\$42.10} in API cost while requiring
  \textbf{36.1 hours of wall-clock runtime} on a single NVIDIA L4 GPU.

% End of appendix - no \end{document} needed as this is included in main.tex

%% file: references.bib
@misc{paischer2024semantic,
      title={Semantic HELM: A Human-Readable Memory for Reinforcement Learning}, 
      author={Fabian Paischer and Thomas Adler and Markus Hofmarcher and Sepp Hochreiter},
      year={2023},
      eprint={2306.09312},
      archivePrefix={arXiv},
      primaryClass={cs.LG},
      url={https://arxiv.org/abs/2306.09312}, 
}

@misc{wang2024state,
      title={LLM-Empowered State Representation for Reinforcement Learning}, 
      author={Boyuan Wang and Yun Qu and Yuhang Jiang and Jianzhun Shao and Chang Liu and Wenming Yang and Xiangyang Ji},
      year={2024},
      eprint={2407.13237},
      archivePrefix={arXiv},
      primaryClass={cs.AI},
      url={https://arxiv.org/abs/2407.13237}, 
}

@misc{chen2023evoprompting,
      title={EvoPrompting: Language Models for Code-Level Neural Architecture Search}, 
      author={Angelica Chen and David M. Dohan and David R. So},
      year={2023},
      eprint={2302.14838},
      archivePrefix={arXiv},
      primaryClass={cs.NE},
      url={https://arxiv.org/abs/2302.14838}, 
}

@misc{wei2025lero,
      title={LERO: LLM-driven Evolutionary framework with Hybrid Rewards and Enhanced Observation for Multi-Agent Reinforcement Learning}, 
      author={Yuan Wei and Xiaohan Shan and Jianmin Li},
      year={2025},
      eprint={2503.21807},
      archivePrefix={arXiv},
      primaryClass={cs.LG},
      url={https://arxiv.org/abs/2503.21807}, 
}

@inproceedings{il1,
author = {Ziebart, Brian D. and Maas, Andrew and Bagnell, J. Andrew and Dey, Anind K.},
title = {Maximum entropy inverse reinforcement learning},
year = {2008},
isbn = {9781577353683},
publisher = {AAAI Press},
booktitle = {Proceedings of the 23rd National Conference on Artificial Intelligence - Volume 3},
pages = {1433–1438},
numpages = {6},
location = {Chicago, Illinois},
series = {AAAI'08}
}

@misc{il3,
      title={Guided Cost Learning: Deep Inverse Optimal Control via Policy Optimization}, 
      author={Chelsea Finn and Sergey Levine and Pieter Abbeel},
      year={2016},
      eprint={1603.00448},
      archivePrefix={arXiv},
      primaryClass={cs.LG},
      url={https://arxiv.org/abs/1603.00448}, 
}

@misc{il4,
      title={Learning Robust Rewards with Adversarial Inverse Reinforcement Learning}, 
      author={Justin Fu and Katie Luo and Sergey Levine},
      year={2018},
      eprint={1710.11248},
      archivePrefix={arXiv},
      primaryClass={cs.LG},
      url={https://arxiv.org/abs/1710.11248}, 
}

@misc{il5,
      title={SEABO: A Simple Search-Based Method for Offline Imitation Learning}, 
      author={Jiafei Lyu and Xiaoteng Ma and Le Wan and Runze Liu and Xiu Li and Zongqing Lu},
      year={2024},
      eprint={2402.03807},
      archivePrefix={arXiv},
      primaryClass={cs.LG},
      url={https://arxiv.org/abs/2402.03807}, 
}

@misc{rlhf1,
      title={A Survey of Reinforcement Learning from Human Feedback}, 
      author={Timo Kaufmann and Paul Weng and Viktor Bengs and Eyke Hüllermeier},
      year={2025},
      eprint={2312.14925},
      archivePrefix={arXiv},
      primaryClass={cs.LG},
      url={https://arxiv.org/abs/2312.14925}, 
}

@misc{rlhf2,
      title={PEARL: Zero-shot Cross-task Preference Alignment and Robust Reward Learning for Robotic Manipulation}, 
      author={Runze Liu and Yali Du and Fengshuo Bai and Jiafei Lyu and Xiu Li},
      year={2024},
      eprint={2306.03615},
      archivePrefix={arXiv},
      primaryClass={cs.LG},
      url={https://arxiv.org/abs/2306.03615}, 
}

@misc{llr1,
      title={Language to Rewards for Robotic Skill Synthesis}, 
      author={Wenhao Yu and Nimrod Gileadi and Chuyuan Fu and Sean Kirmani and Kuang-Huei Lee and Montse Gonzalez Arenas and Hao-Tien Lewis Chiang and Tom Erez and Leonard Hasenclever and Jan Humplik and Brian Ichter and Ted Xiao and Peng Xu and Andy Zeng and Tingnan Zhang and Nicolas Heess and Dorsa Sadigh and Jie Tan and Yuval Tassa and Fei Xia},
      year={2023},
      eprint={2306.08647},
      archivePrefix={arXiv},
      primaryClass={cs.RO},
      url={https://arxiv.org/abs/2306.08647}, 
}

@misc{llr2,
      title={Eureka: Human-Level Reward Design via Coding Large Language Models}, 
      author={Yecheng Jason Ma and William Liang and Guanzhi Wang and De-An Huang and Osbert Bastani and Dinesh Jayaraman and Yuke Zhu and Linxi Fan and Anima Anandkumar},
      year={2024},
      eprint={2310.12931},
      archivePrefix={arXiv},
      primaryClass={cs.RO},
      url={https://arxiv.org/abs/2310.12931}, 
}

@misc{llr3,
      title={DrEureka: Language Model Guided Sim-To-Real Transfer}, 
      author={Yecheng Jason Ma and William Liang and Hung-Ju Wang and Sam Wang and Yuke Zhu and Linxi Fan and Osbert Bastani and Dinesh Jayaraman},
      year={2024},
      eprint={2406.01967},
      archivePrefix={arXiv},
      primaryClass={cs.RO},
      url={https://arxiv.org/abs/2406.01967}, 
}

@InProceedings{pmlr-v235-li24cp,
  title = 	 {{E}vo{R}ainbow: Combining Improvements in Evolutionary Reinforcement Learning for Policy Search},
  author =       {Li, Pengyi and Zheng, Yan and Tang, Hongyao and Fu, Xian and Hao, Jianye},
  booktitle = 	 {Proceedings of the 41st International Conference on Machine Learning},
  pages = 	 {29427--29447},
  year = 	 {2024},
  editor = 	 {Salakhutdinov, Ruslan and Kolter, Zico and Heller, Katherine and Weller, Adrian and Oliver, Nuria and Scarlett, Jonathan and Berkenkamp, Felix},
  volume = 	 {235},
  series = 	 {Proceedings of Machine Learning Research},
  month = 	 {21--27 Jul},
  publisher =    {PMLR},
  url = 	 {https://proceedings.mlr.press/v235/li24cp.html}
}

@misc{hao2023erlre2efficientevolutionaryreinforcement,
      title={ERL-Re$^2$: Efficient Evolutionary Reinforcement Learning with Shared State Representation and Individual Policy Representation}, 
      author={Jianye Hao and Pengyi Li and Hongyao Tang and Yan Zheng and Xian Fu and Zhaopeng Meng},
      year={2023},
      eprint={2210.17375},
      archivePrefix={arXiv},
      primaryClass={cs.NE},
      url={https://arxiv.org/abs/2210.17375}, 
}

@misc{pourchot2019cemrlcombiningevolutionarygradientbased,
      title={CEM-RL: Combining evolutionary and gradient-based methods for policy search}, 
      author={Aloïs Pourchot and Olivier Sigaud},
      year={2019},
      eprint={1810.01222},
      archivePrefix={arXiv},
      primaryClass={cs.LG},
      url={https://arxiv.org/abs/1810.01222}, 
}

@misc{novikov2025alphaevolvecodingagentscientific,
      title={AlphaEvolve: A coding agent for scientific and algorithmic discovery}, 
      author={Alexander Novikov and Ngân Vũ and Marvin Eisenberger and Emilien Dupont and Po-Sen Huang and Adam Zsolt Wagner and Sergey Shirobokov and Borislav Kozlovskii and Francisco J. R. Ruiz and Abbas Mehrabian and M. Pawan Kumar and Abigail See and Swarat Chaudhuri and George Holland and Alex Davies and Sebastian Nowozin and Pushmeet Kohli and Matej Balog},
      year={2025},
      eprint={2506.13131},
      archivePrefix={arXiv},
      primaryClass={cs.AI},
      url={https://arxiv.org/abs/2506.13131}, 
}

@misc{openevolve,
  title = {OpenEvolve: an open-source evolutionary coding agent},
  author = {Asankhaya Sharma},
  year = {2025},
  publisher = {GitHub},
  url = {https://github.com/algorithmicsuperintelligence/openevolve}
}

@misc{brax,
  author = {C. Daniel Freeman and Erik Frey and Anton Raichuk and Sertan Girgin and Igor Mordatch and Olivier Bachem},
  title = {Brax - A Differentiable Physics Engine for Large Scale Rigid Body Simulation},
  url = {http://github.com/google/brax},
  version = {0.14.1},
  year = {2021},
}

@inproceedings{
    xland-minigrid,
    title={{XL}and-MiniGrid: Scalable Meta-Reinforcement Learning Environments in {JAX}},
    author={Alexander Nikulin and Vladislav Kurenkov and Ilya Zisman and Viacheslav Sinii and Artem Agarkov and Sergey Kolesnikov},
    booktitle={Intrinsically-Motivated and Open-Ended Learning Workshop, NeurIPS2023},
    year={2023},
    url={https://openreview.net/forum?id=xALDC4aHGz}
}

@inproceedings{mujoco,
  title={MuJoCo: A physics engine for model-based control},
  author={Todorov, Emanuel and Erez, Tom and Tassa, Yuval},
  booktitle={2012 IEEE/RSJ International Conference on Intelligent Robots and Systems},
  pages={5026--5033},
  year={2012},
  organization={IEEE},
  doi={10.1109/IROS.2012.6386109}
}

@misc{schulman2017proximal,
      title={Proximal Policy Optimization Algorithms}, 
      author={John Schulman and Filip Wolski and Prafulla Dhariwal and Alec Radford and Oleg Klimov},
      year={2017},
      eprint={1707.06347},
      archivePrefix={arXiv},
      primaryClass={cs.LG},
      url={https://arxiv.org/abs/1707.06347}, 
}

@book{sutton2018reinforcement,
  title={Reinforcement Learning: An Introduction},
  author={Sutton, Richard S and Barto, Andrew G},
  year={2018},
  publisher={MIT Press}
}

@book{puterman1994markov,
  title={Markov Decision Processes: Discrete Stochastic Dynamic Programming},
  author={Puterman, Martin L.},
  year={1994},
  publisher={Wiley}
}

@misc{mouret2015illuminatingsearchspacesmapping,
      title={Illuminating search spaces by mapping elites}, 
      author={Jean-Baptiste Mouret and Jeff Clune},
      year={2015},
      eprint={1504.04909},
      archivePrefix={arXiv},
      primaryClass={cs.AI},
      url={https://arxiv.org/abs/1504.04909}, 
}

@article{pugh,
    
AUTHOR={Pugh, Justin K.  and Soros, Lisa B.  and Stanley, Kenneth O. },
           
TITLE={Quality Diversity: A New Frontier for Evolutionary Computation},
          
JOURNAL={Frontiers in Robotics and AI},
          
VOLUME={Volume 3 - 2016},
  
YEAR={2016},
  
URL={https://www.frontiersin.org/journals/robotics-and-ai/articles/10.3389/frobt.2016.00040},
  
DOI={10.3389/frobt.2016.00040},
  
ISSN={2296-9144}}

@misc{austin2021programsynthesislargelanguage,
      title={Program Synthesis with Large Language Models}, 
      author={Jacob Austin and Augustus Odena and Maxwell Nye and Maarten Bosma and Henryk Michalewski and David Dohan and Ellen Jiang and Carrie Cai and Michael Terry and Quoc Le and Charles Sutton},
      year={2021},
      eprint={2108.07732},
      archivePrefix={arXiv},
      primaryClass={cs.PL},
      url={https://arxiv.org/abs/2108.07732}, 
}

@InProceedings{arora2023online,
  title = 	 {Online Inverse Reinforcement Learning with Learned Observation Model},
  author =       {Arora, Saurabh and Doshi, Prashant and Banerjee, Bikramjit},
  booktitle = 	 {Proceedings of The 6th Conference on Robot Learning},
  pages = 	 {1468--1477},
  year = 	 {2023},
  editor = 	 {Liu, Karen and Kulic, Dana and Ichnowski, Jeff},
  volume = 	 {205},
  series = 	 {Proceedings of Machine Learning Research},
  month = 	 {14--18 Dec},
  publisher =    {PMLR},
  url = 	 {https://proceedings.mlr.press/v205/arora23a.html}
}

@inproceedings{levine2010feature,
 author = {Levine, Sergey and Popovic, Zoran and Koltun, Vladlen},
 booktitle = {Advances in Neural Information Processing Systems},
 editor = {J. Lafferty and C. Williams and J. Shawe-Taylor and R. Zemel and A. Culotta},
 pages = {},
 publisher = {Curran Associates, Inc.},
 title = {Feature Construction for Inverse Reinforcement Learning},
 url = {https://proceedings.neurips.cc/paper_files/paper/2010/file/a8f15eda80c50adb0e71943adc8015cf-Paper.pdf},
 volume = {23},
 year = {2010}
}

@article{whitley1999island,
  title={The Island Model Genetic Algorithm: On Separability, Population Size and Convergence},
  author={Whitley, Darrell and Rana, Soraya and Heckendorn, Robert B.},
  journal={Journal of Computing and Information Technology},
  volume={7},
  number={1},
  pages={33--47},
  year={1999}
}
